\documentclass{article} 
\usepackage[final]{colm2026_conference}

\usepackage{microtype}
\usepackage{hyperref}
\usepackage{url}
\usepackage{booktabs}
\usepackage{wrapfig}
\usepackage{multirow}

\usepackage[breakable]{tcolorbox}
\tcbuselibrary{skins}

\definecolor{cyan10}{HTML}{E5F6FF}
\definecolor{cyan20}{HTML}{BAE6FF}
\definecolor{cyan60}{HTML}{0072c3}
\definecolor{cyan70}{HTML}{00539a}
\definecolor{cyan80}{HTML}{003a6d}
\definecolor{teal10}{HTML}{D9FBFB}
\definecolor{teal20}{HTML}{9EF0F0}
\definecolor{teal60}{HTML}{007d79}
\definecolor{orange10}{HTML}{FFF2E8}
\definecolor{orange20}{HTML}{FFD9BE}
\definecolor{orange60}{HTML}{ba4e00}
\definecolor{blue10}{HTML}{EDF5FF}
\definecolor{blue20}{HTML}{D0E2FF}
\definecolor{blue70}{HTML}{0043ce}
\definecolor{blue80}{HTML}{002d9c}
\definecolor{magenta10}{HTML}{FFF0F7}
\definecolor{magenta20}{HTML}{FFD6E8}
\definecolor{magenta30}{HTML}{ffafd2}
\definecolor{magenta50}{HTML}{ee5396}
\definecolor{magenta60}{HTML}{d02670}
\definecolor{magenta70}{HTML}{9f1853}
\definecolor{purple10}{HTML}{F6F2FF}
\definecolor{purple20}{HTML}{E8DAFF}
\definecolor{purple30}{HTML}{d4bbff}
\definecolor{purple70}{HTML}{8a3ffc}
\definecolor{rose10}{HTML}{FCF2ED}
\definecolor{rose20}{HTML}{F9D9D1}
\definecolor{rose60}{HTML}{ab5638}
\definecolor{rose70}{HTML}{853c27}
\definecolor{red10}{HTML}{FFF1F1}
\definecolor{red20}{HTML}{FFD7D9}
\definecolor{green10}{HTML}{DEFBE6}
\definecolor{green20}{HTML}{A7F0BA}
\definecolor{green70}{HTML}{0e6027}
\definecolor{green80}{HTML}{044317}
\definecolor{yellow10}{HTML}{fcf4d6}
\definecolor{yellow20}{HTML}{fddc69}
\definecolor{gray20}{HTML}{e0e0e0}
\definecolor{gray30}{HTML}{c6c6c6}
\definecolor{gray40}{HTML}{a8a8a8}
\definecolor{gray80}{HTML}{393939}

\usepackage{xcolor}
\definecolor{claudecream}{RGB}{250, 247, 242}   
\definecolor{claudecream}{RGB}{253, 251, 247}   
\definecolor{claudecream}{RGB}{250, 247, 242}
\definecolor{claudeborder}{RGB}{140, 128, 110}   

\tcbset{
  aibox/.style={
    width=\linewidth,
    top=7pt,
    bottom=2pt,
    left=2pt,
    right=2pt,
    colback=claudecream,
    colframe=claudeborder,
    colbacktitle=claudeborder,
    enhanced,
    center,
    attach boxed title to top left={yshift=-0.1in,xshift=0.1in},
    boxed title style={boxrule=0pt,colframe=white,},
  }
}

\usepackage{lineno}

\definecolor{darkblue}{rgb}{0, 0, 0.5}
\hypersetup{colorlinks=true, citecolor=darkblue, linkcolor=darkblue, urlcolor=darkblue}

\newtcolorbox{AIbox}[2][]{aibox,title=#2,label=#1}

\usepackage[table, dvipsnames]{xcolor}

\definecolor{codepurple}{rgb}{0.58,0,0.82}
\definecolor{codeblue}{rgb}{0,0,1}
\definecolor{backcolour}{rgb}{0.95,0.95,0.92}

\usepackage{tikz}
\usetikzlibrary{arrows.meta, positioning}

\usepackage{subcaption}
\usepackage{enumitem}

\usepackage{listings}
\usepackage{xcolor}

\usepackage[T1]{fontenc}
\usepackage{textcomp}

\newtheorem{definition}{Definition}

\usepackage{amsmath,amsfonts,bm}

\def\eqref#1{equation~\ref{#1}}

\def\1{\bm{1}}

\DeclareMathAlphabet{\mathsfit}{\encodingdefault}{\sfdefault}{m}{sl}
\SetMathAlphabet{\mathsfit}{bold}{\encodingdefault}{\sfdefault}{bx}{n}

\title{Why Do Reasoning Models Lose Coverage?\\The Role of Data and Forks in the Road}

\author{Ngoc-Hieu Nguyen$^{*}$ \\
The Pennsylvania State University\\
\texttt{hnn5071@psu.edu} \\
\And
Parshin Shojaee\thanks{Equal contribution.}\\
Virginia Tech \\
\AND
Phuc Minh Nguyen \\
VinUniversity \\
\And
Nan Zhang \\
The Pennsylvania State University\\
\And
Chandan K Reddy \\
Virginia Tech \\
\And
Khoa D Doan$^{\dagger}$ \\
VinUniversity \\
\And
Rui Zhang\thanks{Equal advising.} \\
The Pennsylvania State University\\
}

\begin{document}

\ifcolmsubmission
\linenumbers
\fi

\maketitle

\begin{abstract}
\looseness=-1 Recent progress in large language models has led to the emergence of reasoning models, which have shown strong performance on complex tasks through specialized fine-tuning procedures. While these methods reliably improve pass@1 accuracy, prior works have observed that they show a coverage shrinkage behavior, where pass@k degrades relative to the base model.
In this paper, we investigate the cause of reasoning shrinkage under SFT-based post-training. We hypothesize that this behavior is driven by properties of the fine-tuning data, specifically related to decision points or ``forks in the road'' scenarios where model encounters indecipherable patterns with multiple valid reasoning paths.
To test this hypothesis, we design controlled case studies that simulate such decision-point settings, spanning indecipherable nodes in graph branching, and reasoning modes. By tracking post-training dynamics in these settings, we find that the shrinkage phenomenon is tightly correlated with the prevalence of decision-point scenarios in the training data.
We also demonstrate that this shrinkage behavior can be partially mitigated through targeted data synthesis design of decision-points, and a more systematic diversity-encouraging decoding mechanism. 
Our findings identify data-centric factors as a key driver of shrinkage in reasoning models and highlight diversity-aware designs as an effective lever for controlling it.\footnote{Data and code for reproducing our experiments are available at: \url{https://github.com/psunlpgroup/reasoning_forks}}
\end{abstract}

\begin{flushright}
\textit{``For every path you choose, there is another you must abandon''}\\
\textit{— Joan D. Vinge}
\end{flushright}

\vspace{-0.5em}
\section{Introduction}
\label{sec:intro}
\vspace{-0.5em}
\looseness=-1 Large language models (LLMs) fine-tuned for reasoning have recently achieved remarkable gains on complex reasoning benchmarks. Through post-training procedures such as reasoning-based supervised fine-tuning (SFT) and reinforcement learning with verification feedback (RLVR) \citep{lambert2025tulu,guo2025r1}, these models reliably have shown improvement on pass@1 accuracy, often substantially outperforming their base counterparts. As a result, reasoning-oriented post-training has become a central paradigm for improving the capabilities of LLMs on challenging tasks.

\looseness=-1 Despite these gains, a growing body of work in literature has identified a concerning and counterintuitive phenomenon known as \emph{coverage shrinkage}, where improvements in pass@1 are accompanied by a degradation in pass@k, indicating a loss of diversity and coverage in the model’s learned reasoning procedures.
Understanding why this happens, and what fundamentally drives this shrinkage, remains an open and important research question.
Most prior works have attributed this behavior to the optimization dynamics of RLVR~\citep{yue2025does,wu2025invisible,liu2025prorl,shojaee2025illusion}, arguing that reward-driven training overemphasizes a narrow set of high-reward reasoning paths and leads to an increased chance of mode collapse.
However, recent works have shown that coverage shrinkage is not unique to RLVR and also arises under supervised post-training with SFT~\citep{chen2025rethinking, dang2025weight}, which implies that shrinkage may not simply be a consequence of post-training algorithms, but may arise from a more fundamental driver.

\begin{wrapfigure}{r}{0.55\textwidth}
\vspace{-5pt}
\centering\includegraphics[width=1.0\linewidth]{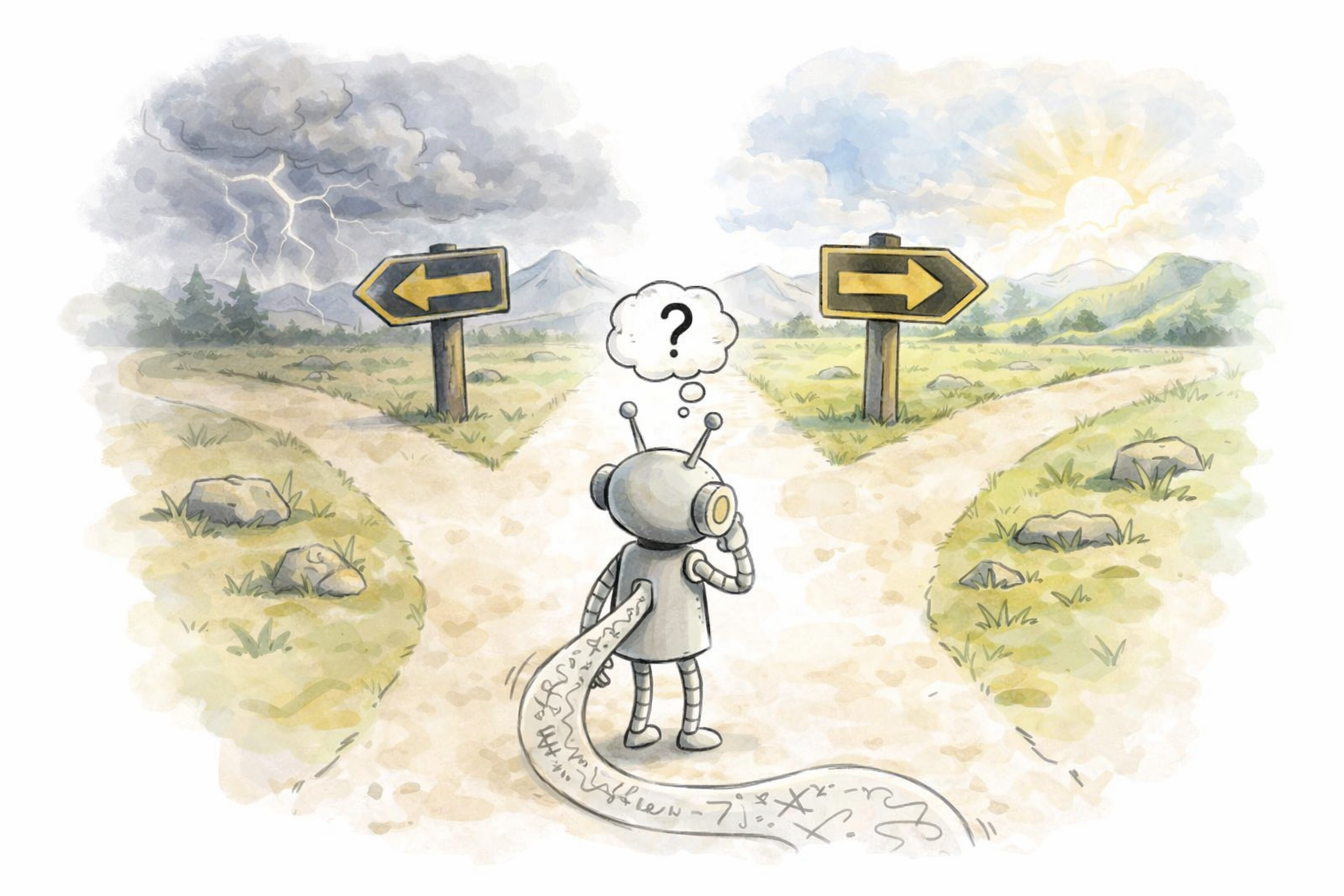}
\caption{
\small \looseness=-1 Reasoning unfolds through \textit{forks in the road}, choices made without knowing the path to truth.}
\label{fig: motiv}
\end{wrapfigure}
\looseness=-1 In this paper, we argue that data plays a central and underexplored role in driving the coverage shrinkage behavior. Instead of focusing on post-training algorithms, we shift attention to the structure and properties of the post-training data, and ask a simple but critical question: \emph{what aspects of reasoning data encourage the models to collapse onto fewer solution paths?}
Our key hypothesis is that shrinkage is strongly influenced by \emph{decision points} or \emph{``forks in the road''} in the data, situations in which a model encounters multiple valid, indistinguishable reasoning paths and must commit to one. 
At such points, post-training implicitly pressures the model to commit to a subset of these options available. Over time, this commitment suppresses alternative trajectories, resulting in improved correctness along dominant paths but reduced overall coverage.

\begin{figure}[ht]
\vspace{-3.0em}
\centering
\includegraphics[width=0.95\linewidth]{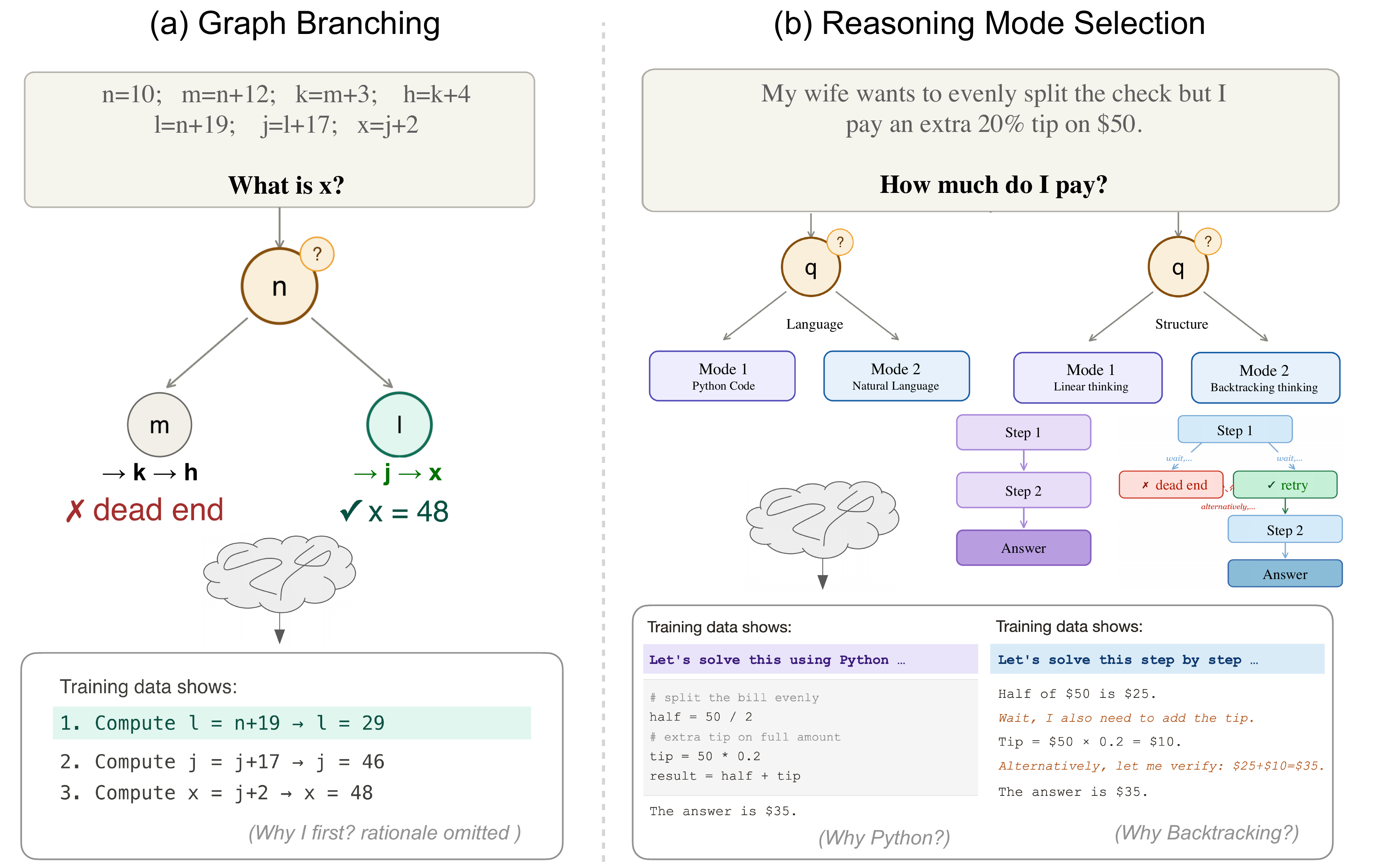}
\caption{\looseness=-1 \small \textbf{Illustrative examples of \textit{forks in the road} case studies.} \textbf{(a)} Graph navigation with indecipherable nodes; and \textbf{(b)} Mathematical reasoning with multiple valid solution modes. In both settings, decision points force commitment to a path without knowing which will succeed.
\vspace{-1em}
}
\label{fig:forks-in-road}
\end{figure}

\looseness=-1 To test this hypothesis, we design controlled case studies that isolate and expose these decision-point structures to the reasoning (Figure~\ref{fig:forks-in-road}).
Our first setting is a graph-based navigation task, a natural testbed for studying decision points, which is inspired by prior work on \emph{indecipherable nodes} in next-token prediction~\citep{bachmann2024the}. In this task, a model must traverse a star graph from a start node to a target node while encountering the branching points that provide no information about which branch leads to success. 
Our second setting focuses on mathematical reasoning problems that admit multiple valid reasoning modes and solution strategies.
At certain stages of reasoning, the model must decide how to proceed without knowing which strategy will ultimately succeed; these moments constitute decision points analogous to the graph branching. 
By tracking post-training dynamics across both of these settings as well as across ablated data variants, we observe that post-training coverage shrinkage is largely driven by how models resolve ambiguity under repeated exposure to such decision points. In particular, ablating or restructuring these points significantly alters model behavior, and the degree of shrinkage is correlated with their prevalence in the post-training data.

\looseness=-1 Motivated by our findings, we introduce two practical strategies to control such shrinkage phenomenon. First, we show that per-problem coverage of alternative decisions can significantly control shrinkage behavior compared to distributing diversity across the problems.
This highlights the importance of synthetic data design that accounts for decision-point structure and its coverage
within problems. Second, we observe that shrinkage is closely associated with the emergence of high-frequency tokens that often correspond to early, indecipherable choices for the model in the reasoning process.
We demonstrate that a simple decoding mechanism that encourages diversity among these dominant initial tokens can partially recover the lost coverage after the post-training without the need for any additional re-training.

\looseness=-1 
This study is mainly aimed to deepen our understanding of coverage shrinkage in reasoning models and provides a data-centric perspective on this phenomenon.
Our key contributions are as follows:
\begin{itemize}[leftmargin=*,itemsep=0pt, topsep=0pt]
\item \looseness=-1 We present a systematic, \emph{data-centric} study of coverage shrinkage in reasoning post-trained models, aiming to understand its underlying factors.

\item \looseness=-1 We identify forks-in-the-road patterns in fine-tuning data as a key driver of coverage shrinkage, and analyze this effect through targeted case studies such as graph branching and alternative mathematical reasoning strategies.

\item Through controlled experiments and training-dynamics analysis, we find a strong correlation between the structure of decision points in data and the severity of coverage shrinkage, providing empirical evidence for the role of data on such behavior.

\item Motivated by our findings, we introduce two simple diversity-aware data synthesis and decoding strategies, and present proof-of-concept results demonstrating their effectiveness in mitigating shrinkage. These results suggest that the lost coverage is not permanently forgotten, but instead suppressed, and can be recovered through effective inference-time intervention.
\end{itemize}

\section{Preliminaries}
\label{sec:prelim}
\vspace{-0.5em}
To formally ground our investigation, we first define the standard metrics and review the mechanics of post-training LLMs.

\paragraph{Coverage Definition.} \looseness=-1 In this study, coverage measures the effectiveness of test-time scaling through repeated sampling~\citep{brown2025large}. 
Let $\pi(\cdot\mid x)$ denote the model’s output distribution given $x$.
For a problem $x$, pass@$k$ is the probability that at least one of $k$ independent samples from $\pi(\cdot|x)$ is correct.
Given a set of prompt $\mathcal{D}_X$ and $n\ge k$ samples per problem, it is estimated as
$$
\operatorname{pass@}k
=\frac{1}{|\mathcal{D}_X|}\sum_{x_i\in\mathcal{D}_X}
\left[1-\frac{\binom{n-c_i}{k}}{\binom{n}{k}}\right],
$$
where $c_i$ is the number of correct samples for $x_i$. 
In the setting of reasoning models \citep{guo2025r1, jaech2024openai}, we sample $k$ thinking chains $\tau_i^{(k)} \sim \pi(\cdot | x_i)$ and corresponding final answers $y_i^{(k)} \sim \pi(\cdot | \tau_i^{(k)}, x_i)$. These chains may differ in phrasing and format \citep{zhao2025echo} or in the reasoning skills they invoke \citep{he2025skill}, giving rise to exploration of reasoning at test-time.

\paragraph{Post-training and Coverage Shrinkage.}
\looseness=-1 In practice, to enhance the reasoning capabilities of LLMs, a common approach is to first perform reasoning-based fine-tuning with SFT on a dataset of triples $(x_i, \tau_i, y_i)$, where $x_i$ denotes a problem instance, $\tau_i$ is a corresponding thinking chain, and $y_i$ is the final answer. For most problems $x_i$, the dataset contains only one or a small number of annotated thinking chains $\tau_i$. As a result, SFT encourages the model to concentrate probability mass on this limited set by maximizing $\pi_\theta(y_i, \tau_i \mid x_i)$. 
This training procedure typically improves pass@1 by increasing the likelihood of generating reasoning chains that closely match the annotated $\tau_i$ and $y_i$ for each $x_i$ in the training set. 
Following SFT, the reinforcement learning with verifiable rewards (RLVR) \citep{lambert2025tulu,guo2025r1} is often applied as another post-training stage, and has demonstrated substantial empirical gains on verifiable reasoning tasks, including mathematics and coding.
Recent analyses have examined both the strengths and limitations of these post-training stages for reasoning, identifying a counterintuitive phenomenon termed as \emph{sharpening} or \emph{coverage shrinkage} where improvements in pass@1 often coincide with declines in pass@$k$ \citep{dang2025weight, chen2025rethinking, yue2025does, wu2025invisible}. This effect is also commonly attributed to reduced diversity of the model’s reasoning paths. In our paper, we primarily seek to understand this coverage shrinkage behavior and provide a data-centric perspective on its underlying factors.

\section{A Data-Centric View: Forks in the Road}
\label{sec:hypothesis}
\vspace{-1.0em}

\looseness=-1 Solving complex reasoning problems is inherently exploratory: a solver often encounters \textit{decision points} along a reasoning trajectory, where multiple valid continuations are possible—such as alternative reasoning branches, modes, or strategies (Figure~\ref{fig:forks-in-road})—without knowing in advance which will succeed. 
Strong problem solvers must therefore consider multiple candidate strategies or continuations.
\looseness=-1 However, current reasoning-focused fine-tuning datasets often suffer from survivorship bias: models are typically exposed only to one or a few successful reasoning paths during post-training. 
Moreover, the rationale behind the choice of the given reasoning path over the alternative strategies is hidden \citep{gandhi2024stream, lehnert2024beyond}.
We refer to decision points with such missing selection rationale as \textit{fork-in-the-road patterns}, informally defined as follows:
\vspace{-0.5em}
\begin{definition}[Fork-in-the-road patterns in reasoning traces]
\label{def:forks-in-the-road}
    A fork-in-the-road pattern is a decision point with multiple viable continuations where the preceding context does not contain sufficient information for models to learn to uniquely determine a single path.
\end{definition}
\vspace{-0.5em}
\looseness=-1 Such patterns can occur at both micro-level (step-by-step or which algebraic manipulation to apply next) and macro-level (strategy or mode selection).
For instance, a model may need to choose between solving a problem through analytical natural-language reasoning or by directly writing a Python program.
When trained to imitate a single continuation without its selection rationale, the model may instead learn \textit{spurious cues} that favor particular reasoning paths or modes.

\looseness=-1 Synthesizing these observations and our empirical results, we propose the following hypothesis regarding the failure modes of reasoning models: 

\begin{AIbox}[]{Fork-in-the-road Hypothesis}{}
\vspace{0.2em}
\small When training data contains \textit{fork-in-the-road patterns}—decision points with multiple viable paths but insufficient rationale for selecting among them—the model fails to learn a generalizable path-selection mechanism.  Instead, it may rely on \textit{spurious, non-causal cues}, causing its output distribution to concentrate on a subset of available reasoning paths. This suppresses alternative trajectories and leads to coverage shrinkage at inference time, reflected by degraded $pass@k$ after post-training.
\vspace{0.2em}
\end{AIbox}

\looseness=-1 To test this hypothesis, we design controlled case studies, as shown in Figure~\ref{fig:forks-in-road}, that isolate and highlight these decision-point structures within the SFT data, as detailed in the following sections.

\section{Experimental Framework}

\subsection{Graph Navigation}
\label{sec:exp-graph}
\vspace{-0.7em}

\looseness=-1 
To enable controlled evaluation of our hypothesis, we begin with a simple synthetic setting that isolates the role of decision points in reasoning processes.
Inspired by \cite{zhang2022unveiling} and \cite{bachmann2024the}, we first focus experiments on a dependency chain evaluation task with a star-graph structure, a natural testbed for decision points.
Concretely, each prompt specifies functional dependencies among variables (graph nodes) and a query over a target variable. Figure~\ref{fig:forks-in-road} (left) shows an example with its ground-truth solution. 
To isolate the role of decision points in coverage shrinkage analysis, we construct two controlled data variants from the same synthetic task that differ only in whether fork-in-the-road patterns are present.

\looseness=-1 
In the \emph{Forward} setting, the ground-truth solutions are presented step-by-step from the root to the target node.
Due to the binary branching nodes, the first step (highlighted in green) is a decision point where both $m$ and $l$ are valid continuations; however, only $l$ leads to the correct target $x$. 
While this can be determined via search, the model is only exposed to the successful trajectory at training, leaving the underlying decision process implicit. This setting matches our definition of the fork-in-the-road pattern (Definition \ref{def:forks-in-the-road}).
In contrast, the \emph{Reverse (w/o DP)} setting removes these forks by presenting the correct trajectory in reverse format (target to source), effectively eliminating any need for path selection by the model.
Because both settings correspond to the same underlying task, any behavioral differences can be directly attributed to the impact of the fork-in-the-road pattern in the data. 
We evaluate this setup on two backbone models (\texttt{Qwen-2.5-0.5B} \citep{qwen2.5} and \texttt{EvoLM-1B} \citep{qi2025evolm}) to test robustness of observations with respect to pre-training initialization prior to the post-training.

\begin{figure}[h]
\centering
\includegraphics[width=1.0\linewidth]{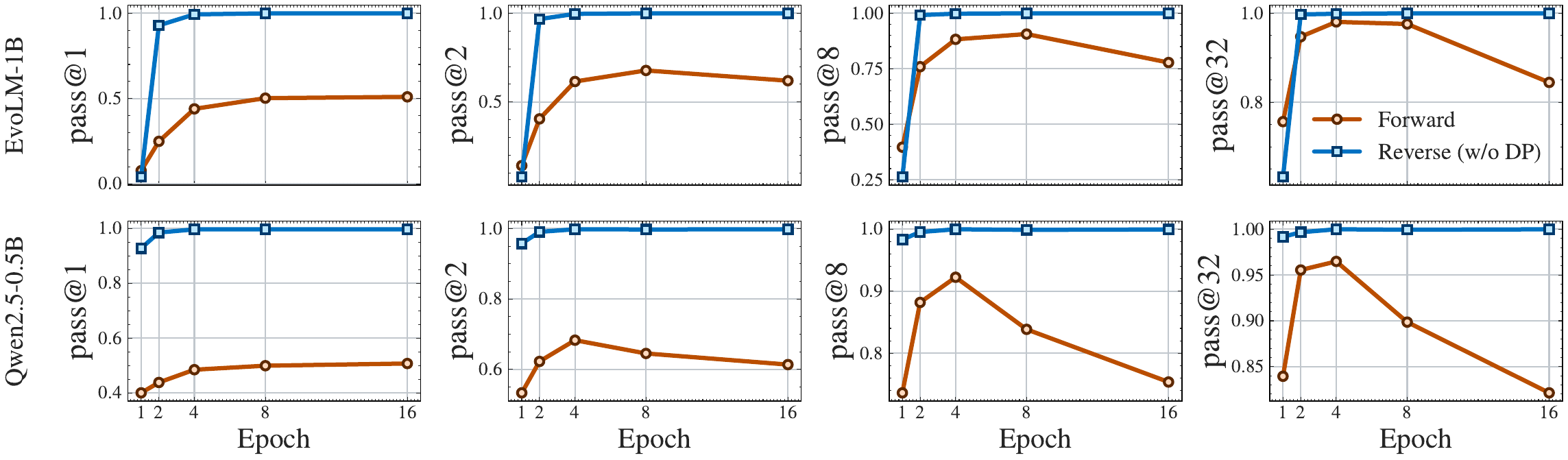}
\vspace{-0.5em}
\caption{
\small \looseness=-1 \textbf{Effect of decision points on coverage in graph navigation task.} Pass@k across SFT epochs for \textcolor{rose60}{\emph{Forward}} vs. \textcolor{cyan70}{\emph{Reverse} (w/o DP)} problem solving settings.
}
\label{fig:graph-passk}
\end{figure}

\vspace{-0.5em}
\paragraph{Results.}
\looseness=-1 Figure~\ref{fig:graph-passk} shows a clear divergence between these two settings.
When decision points are removed (\textcolor{cyan70}{\emph{Reverse (w/o DP)}}), both models quickly reach near-perfect $pass@k$ performance across all values of $k$, and this performance remains stable even for large $k$ and extended training time. 
However, when decision points are present (\textcolor{rose60}{\emph{Forward}}), the behavior completely changes. While pass@1 improves steadily, performance for larger $k$ follows a different pattern: it first improves, then degrades as training continues. This degradation becomes more pronounced as $k$ increases, providing direct evidence of coverage shrinkage behavior.

\begin{wrapfigure}{r}{0.5\textwidth} 
\centering
\vspace{-1.0em}
\begin{subfigure}[t]{0.49\linewidth}
\centering
\includegraphics[width=\linewidth]{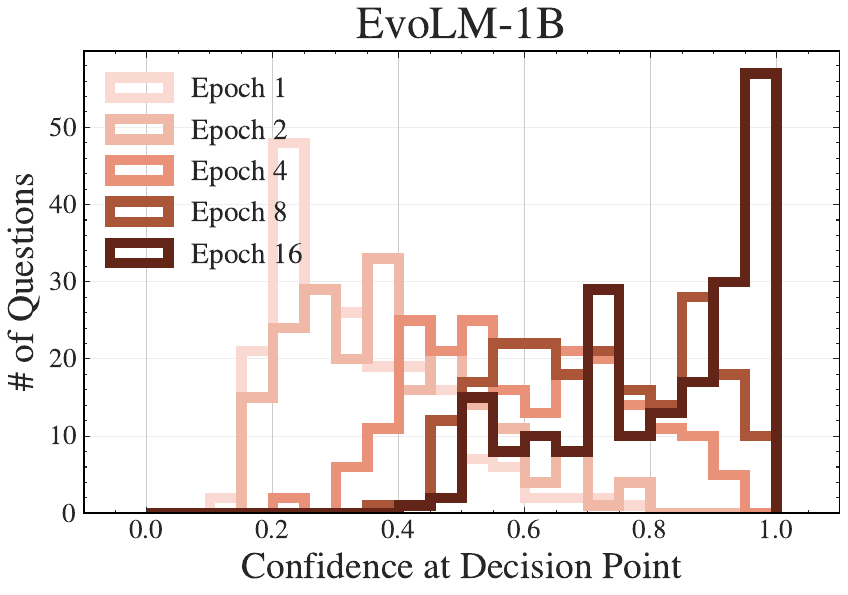}
\end{subfigure}
\hfill
\begin{subfigure}[t]{0.49\linewidth}
\centering
\includegraphics[width=\linewidth]{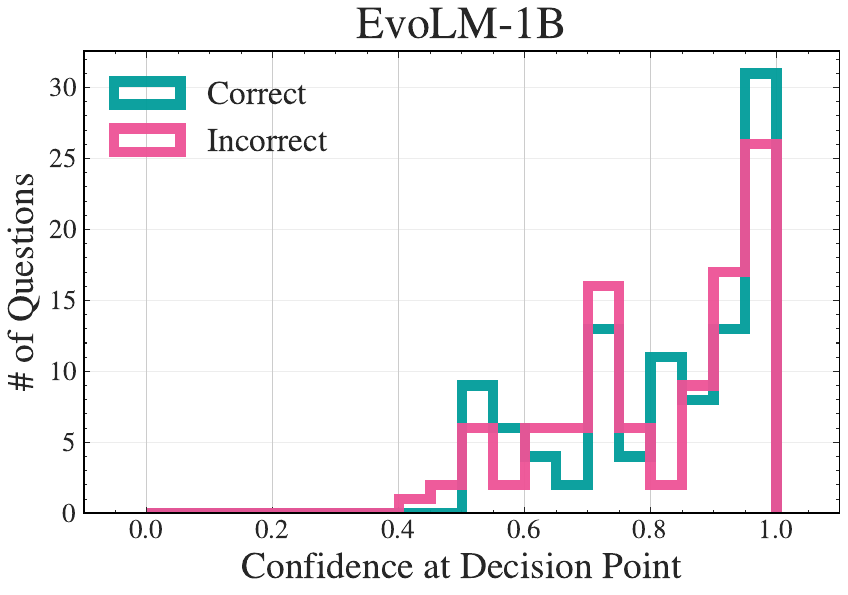}
\end{subfigure}

\vspace{0.5em}

\begin{subfigure}[t]{0.49\linewidth}
\centering
\includegraphics[width=\linewidth]{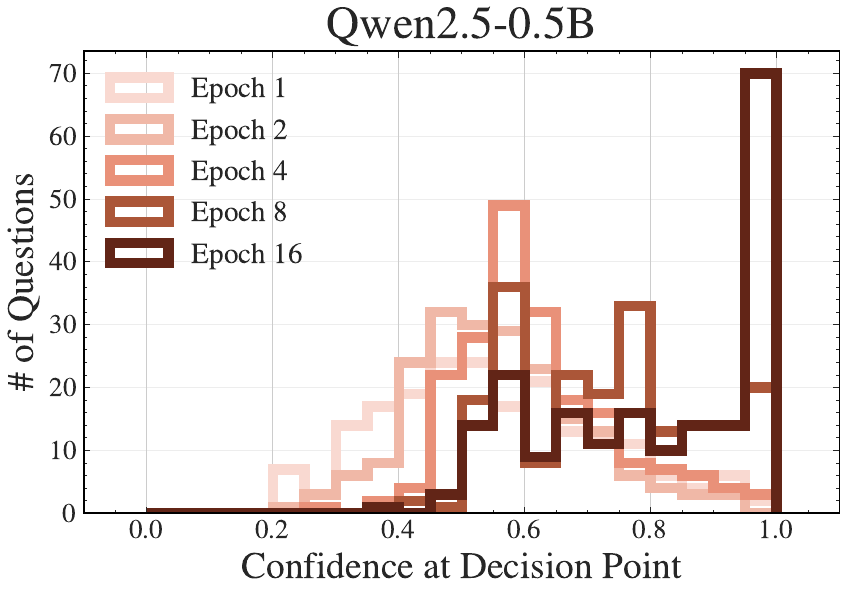}
\end{subfigure}
\hfill
\begin{subfigure}[t]{0.49\linewidth}
\centering
\includegraphics[width=\linewidth]{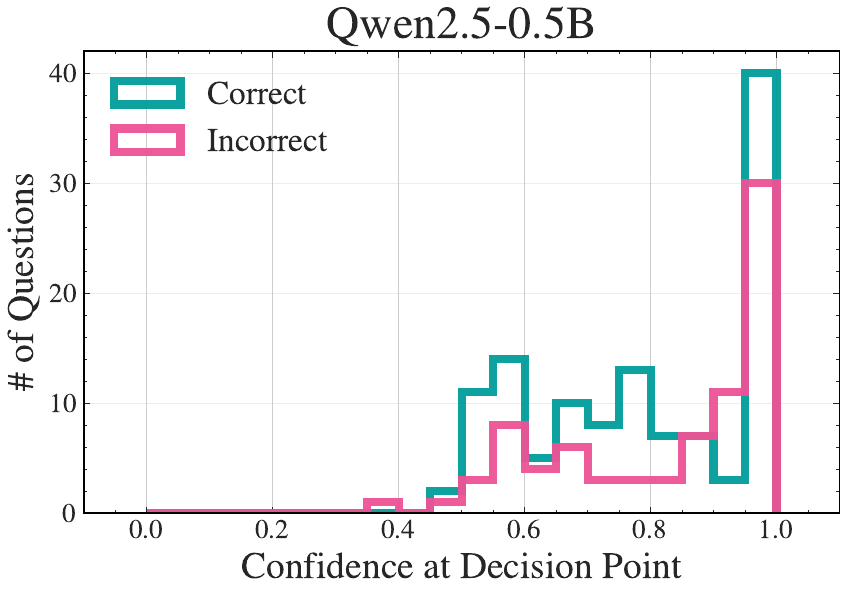}
\end{subfigure}
\caption{
\small \looseness=-1 \textbf{Left:} Change in model confidence at decision points over the course of SFT; 
\textbf{Right:} At the last epoch, the model assigns high confidence to both correct and incorrect paths, indicating uncalibrated decisions that drive coverage collapse.
\vspace{-5em}
}
\label{fig:graph-conf}
\end{wrapfigure}

\looseness=-1 We also looked deeper into \textcolor{rose60}{\emph{Forward}} model behavior at decision points (Figure~\ref{fig:graph-conf}). In this setting, we observe that the model’s confidence at decision points (measured by token-level probability) increases sharply throughout the epochs of training (\textbf{left}). However, this increase is not selective: the model is highly confident not only when it chooses the correct branch, but also when it chooses an incorrect one (\textbf{right}). This shows that training with fork-in-the-road patterns in data can push the model toward overconfident, single-path commitments, rather than calibrated uncertainty of correct solutions over multiple valid continuations. 

Additional analysis in Appendix~\ref{sec:appendix_spurious_cues_graph} shows that branch selection is fragile to semantically equivalent prompt perturbations, indicating reliance on spurious cues at decision points.
As a result, alternative trajectories are progressively suppressed, leading to the observed coverage shrinkage and drop in pass@k. 

\looseness=-1 We also validate our hypothesis under RLVR (Appendix~\ref{sec:grpo_graph}): the same coverage shrinkage emerges in the forward but not reverse setting, suggesting that it is driven not only by the learning algorithm but also by data structure and decision points.

\vspace{3.0em}
\subsection{Reasoning Mode Selection}
\label{sec:exp-reason}

\looseness=-1 The forks-in-the-road phenomenon is not limited to synthetic graph navigation settings; it also arises naturally in real-world reasoning tasks where multiple high-level solution strategies coexist. 
During generation, the model must implicitly commit to one strategy early, before knowing which will succeed. These early commitments act as decision points, analogous to branching in graph-based settings.
To study this effect, we examine reasoning mode selection as a representative instance of such decision points. Specifically, we consider two settings: (i) different reasoning representations: natural language (NL) versus code-based reasoning, and (ii) different reasoning structures: linear vs. backtracking reasoning. We analyze each of these settings in detail below.

\begin{figure}[b]
\centering
\includegraphics[width=\linewidth]{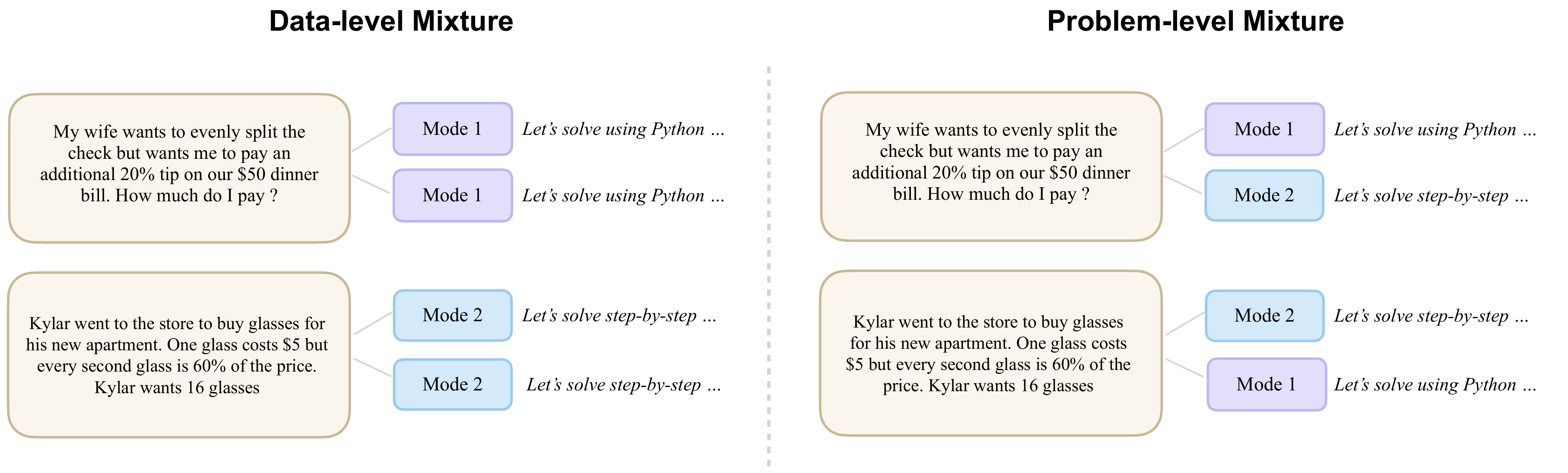}
\caption{\looseness=-1 \small \textbf{Diversity structure in SFT data.} Data-level diversity distributes modes across problems, while problem-level diversity exposes multiple reasoning modes within each problem.}
\label{fig:reason-datadiv}
\end{figure}

\subsubsection{Natural Language vs. Code Reasoning}
\label{sec:exp-reason-lang}

\looseness=-1 We treat the choice between natural language (NL) and code-based reasoning as a decision point (following Definition \ref{def:forks-in-the-road}). 
In this section, we investigate whether models trained on mixed data can learn to balance different reasoning modes under repeated sampling.
A key question is how the \emph{structure} of diversity in training data affects this decision. In our experiments, we construct two data designs with identical diversity ratios (50\% NL, 50\% code) but different organization (Figure~\ref{fig:reason-datadiv}): \textbf{Data-level diversity:} each problem is solved using a single mode, but the dataset is globally balanced across the modes; \textbf{Problem-level diversity:} each problem appears in fine-tuning data with both reasoning modes.
This setup helps us to better understand whether coverage depends on just \emph{how much} diversity is present, or  \emph{how it is also distributed}.

\vspace{-0.7em}
\paragraph{Setup.} 
\looseness=-1 In this analysis, we collect SFT samples from two common mathematical reasoning datasets:  
OpenMathInstruct-1 \citep{toshniwal2024openmath}, and OpenMathInstruct-2 \citep{toshniwal2024openmath2}. 
The OpenMathInstruct-2 consist of solutions in natural language, while OpenMathInstruct-1 consists of Python code solutions generated by Mixtral-8x7B \citep{jiang2024mixtral}. 
We first filter only GSM8k questions and combine them from these two datasets.
We sample the solution such that each reasoning style (NL vs Code) is distributed equally in the whole fine-tuning dataset.
Therefore, any inherent differences in dataset quality or source are held constant across the comparison.
For the evaluation phase, we use questions from \textbf{the test set} of the original GSM8k dataset.

\begin{figure}[h]
\centering
\includegraphics[width=\textwidth]{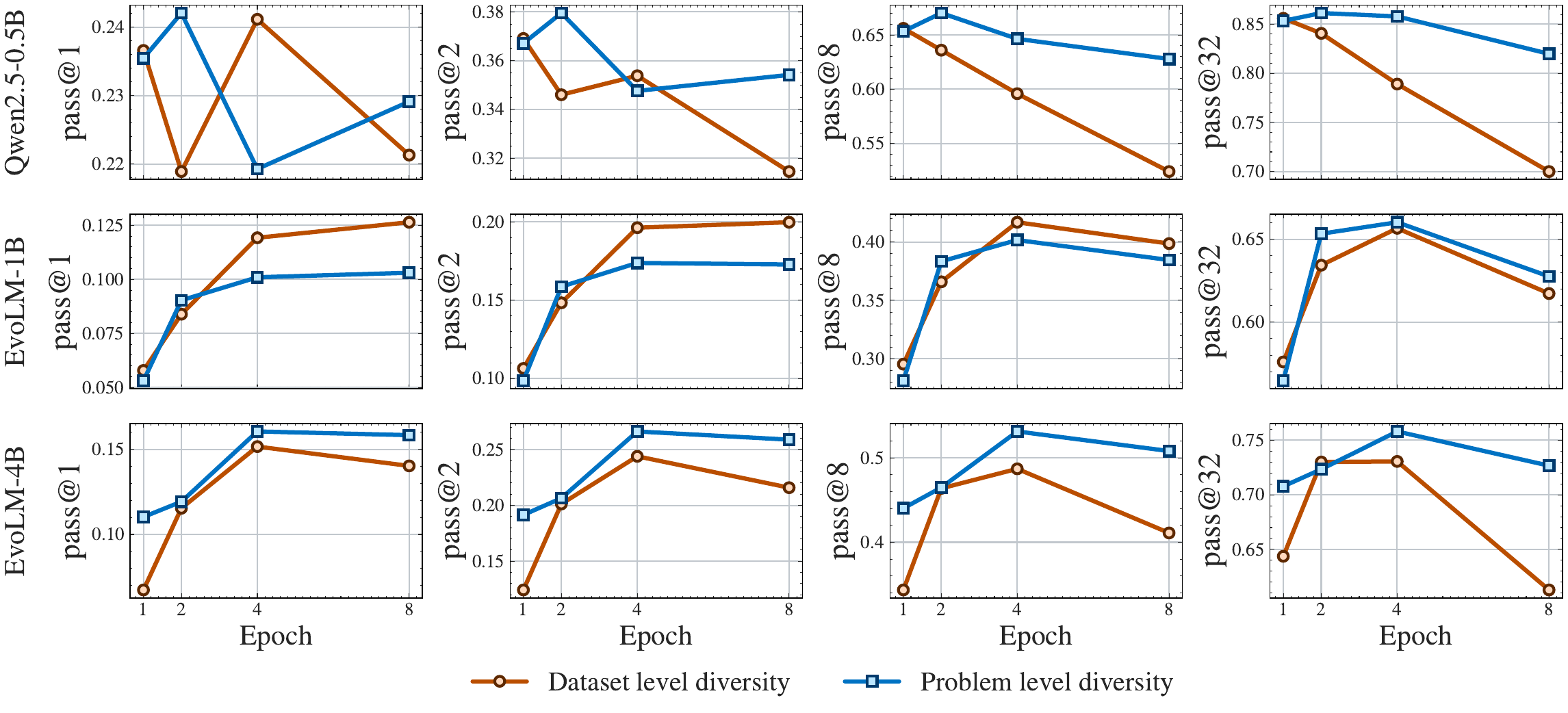}
\caption{
\small \textbf{Effect of data diversity design on coverage in reasoning mode selection task}. Pass@k evaluated on the GSM8K test split across SFT epochs for \textcolor{rose60}{\emph{data-level}} vs. \textcolor{cyan70}{\emph{problem-level}} data diversity.}
\label{fig:reason-passk}
\end{figure}

\paragraph{Results.} 
\looseness=-1 Figure~\ref{fig:reason-passk} highlights a key point that is easy to miss when considering only overall data diversity: even when the ratio between reasoning modes is the same, varying only how problems are solved leads to significantly different coverage behavior after post-training.
Under \textcolor{cyan70}{\emph{problem-level diversity}}, performance is stable across epochs, with only a small drop in pass@k at larger $k$. By contrast, \textcolor{rose60}{\emph{data-level diversity}} exhibits the opposite pattern where pass@1 improves monotonically, but pass@k for larger $k$ initially increases and then degrades sharply as training continues. This effect is most visible at high $k$, where the reduction in pass@k becomes substantial.
Importantly, these differences arise despite both settings having the same overall diversity ratio. The only change is in how that diversity is distributed. 
This suggests that coverage is not determined solely by how much diversity is present; it depends critically on how that diversity is also structured between problems.

\begin{figure}[t]
\centering
\begin{minipage}{\textwidth}
\centering
\begin{subfigure}{\linewidth}
\centering
\includegraphics[width=0.32\linewidth]{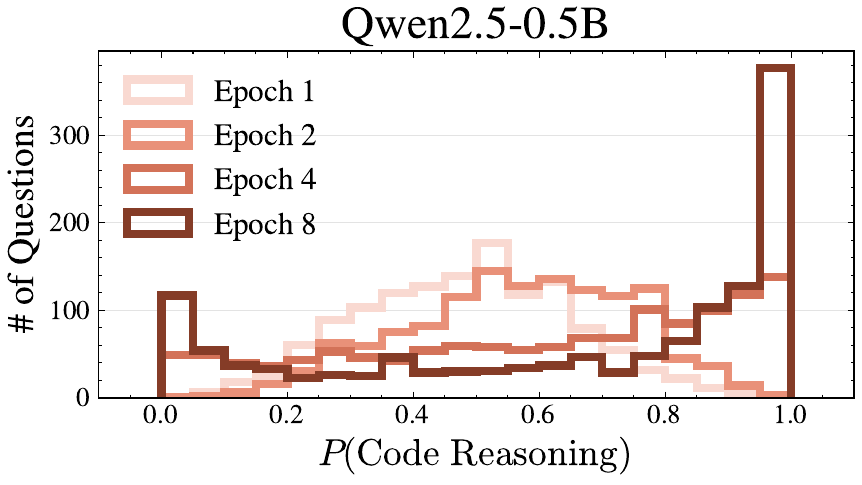}
\includegraphics[width=0.32\linewidth]{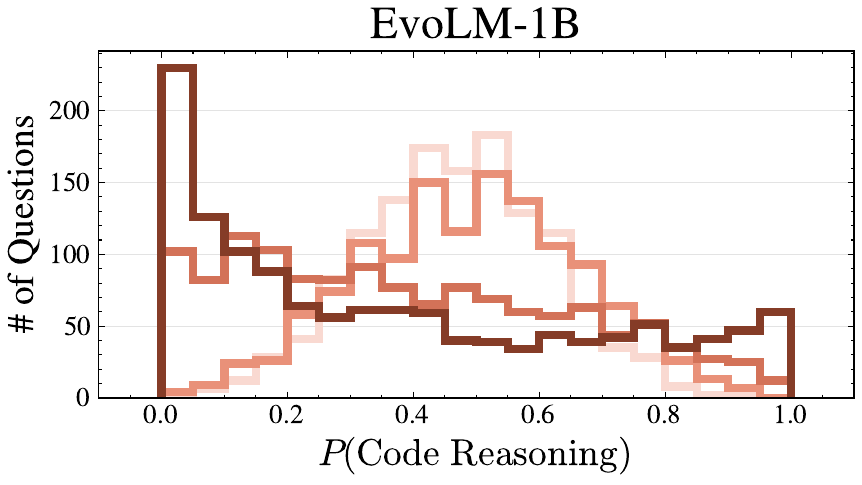}
\includegraphics[width=0.32\linewidth]{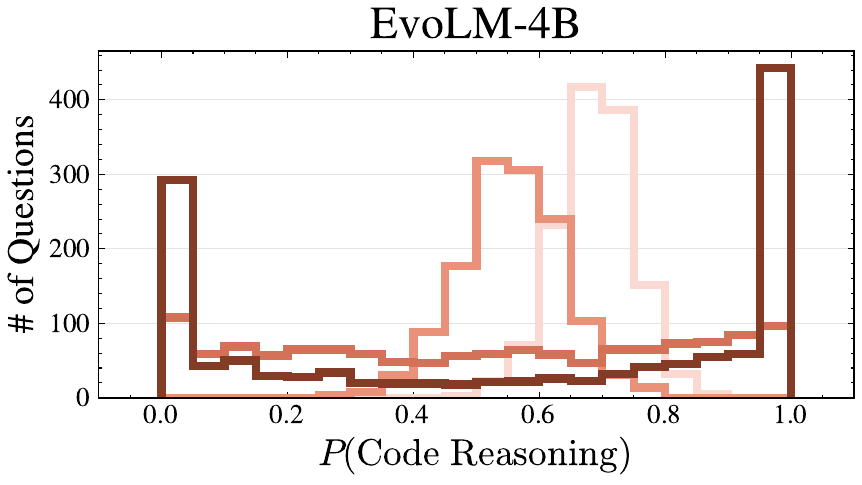}
\vspace{-0.2em}
\caption{Dataset level diversity}
\end{subfigure}

\vspace{0.5em}

\begin{subfigure}{\linewidth}
\centering
\includegraphics[width=0.32\linewidth]{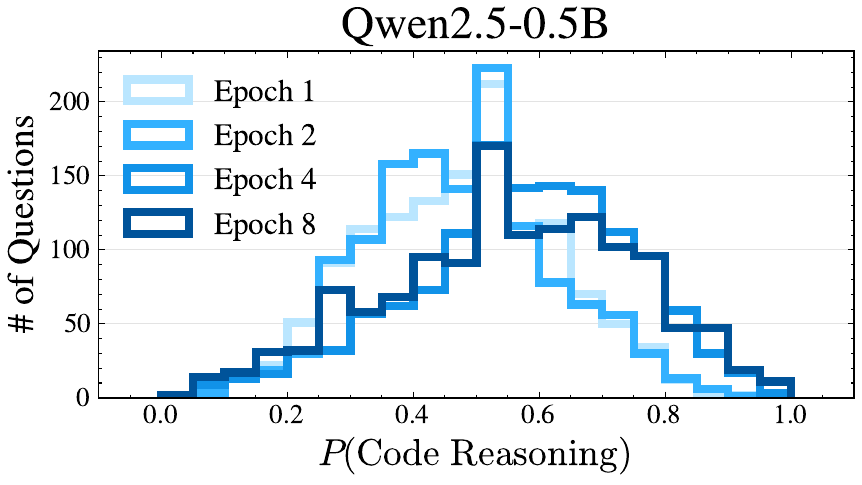}
\includegraphics[width=0.32\linewidth]{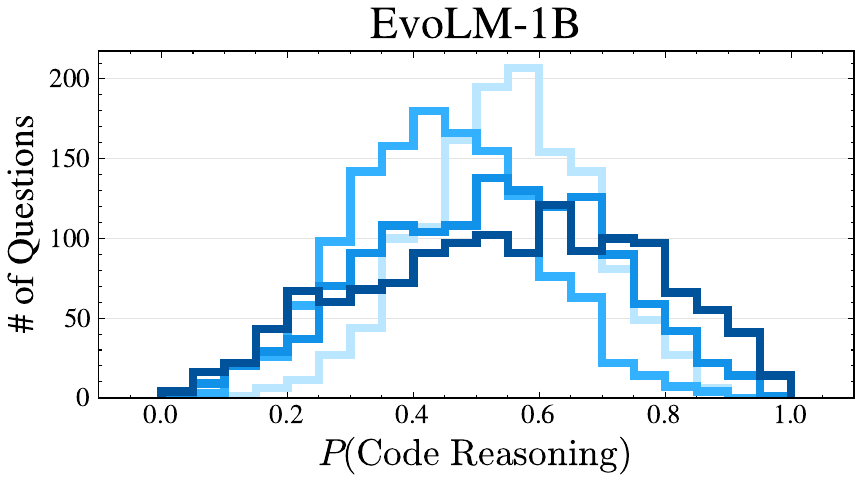}
\includegraphics[width=0.32\linewidth]{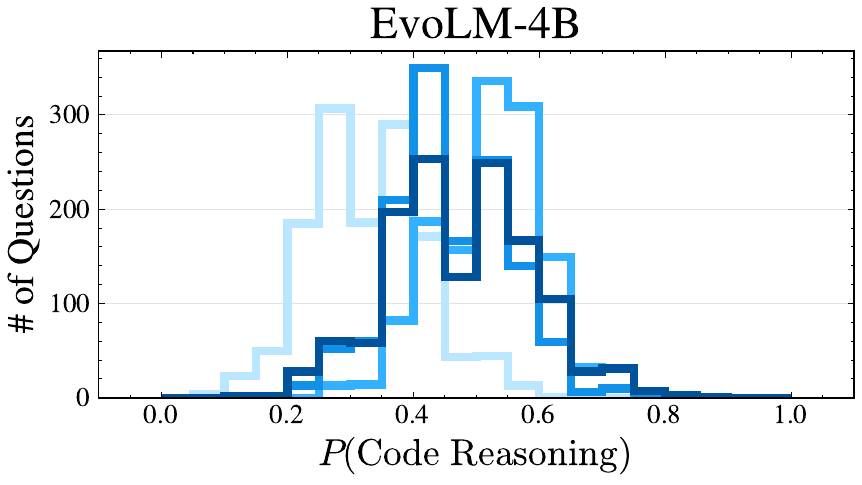}
\vspace{-0.2em}
\caption{Problem level diversity}
\end{subfigure}
\end{minipage}
\caption{\looseness=-1 \small \textbf{Reasoning mode preference under different diversity designs.}
Distribution of per problem code vs. NL reasoning across SFT epochs  for \textcolor{rose60}{\emph{data-level}} vs. \textcolor{cyan70}{\emph{problem-level}} diversity.}
\label{fig:reason-conf}
\end{figure}

\looseness=-1 To better understand this difference, we also analyze the model’s preference over reasoning modes by measuring the probability of generating code vs. NL solutions for each question (as shown in Figure~\ref{fig:reason-conf}).
Under {\emph{data-level diversity}}, the model becomes increasingly confident in selecting a mode per problem, leading to a bimodal distribution that favors either code or NL (\textbf{top}).
This aligns with the overconfidence at decision points in the graph setting, where increasing certainty concentrates probability mass on a few trajectories, causing coverage shrinkage and a drop in pass@k.
In contrast, {\emph{problem-level diversity}} yields a completely different, more balanced, and calibrated distribution (\textbf{bottom}). Rather than committing to a path, it preserves flexibility, which helps maintain coverage across different solution paths.

These observations suggest models must infer which reasoning mode to apply despite an implicit selection mechanism, often driven by annotator preferences or data curation. Mixing styles can thus introduce bias. As we show next, analogous to the fork-in-the-road step, this leads models to rely on spurious features when selecting reasoning modes.

\subsubsection{Linear vs. Backtracking Reasoning}
\label{sec:exp-reason-structure}

\begin{figure}[b]
\centering
\vspace{-1.5em}
\includegraphics[width=0.85\linewidth]{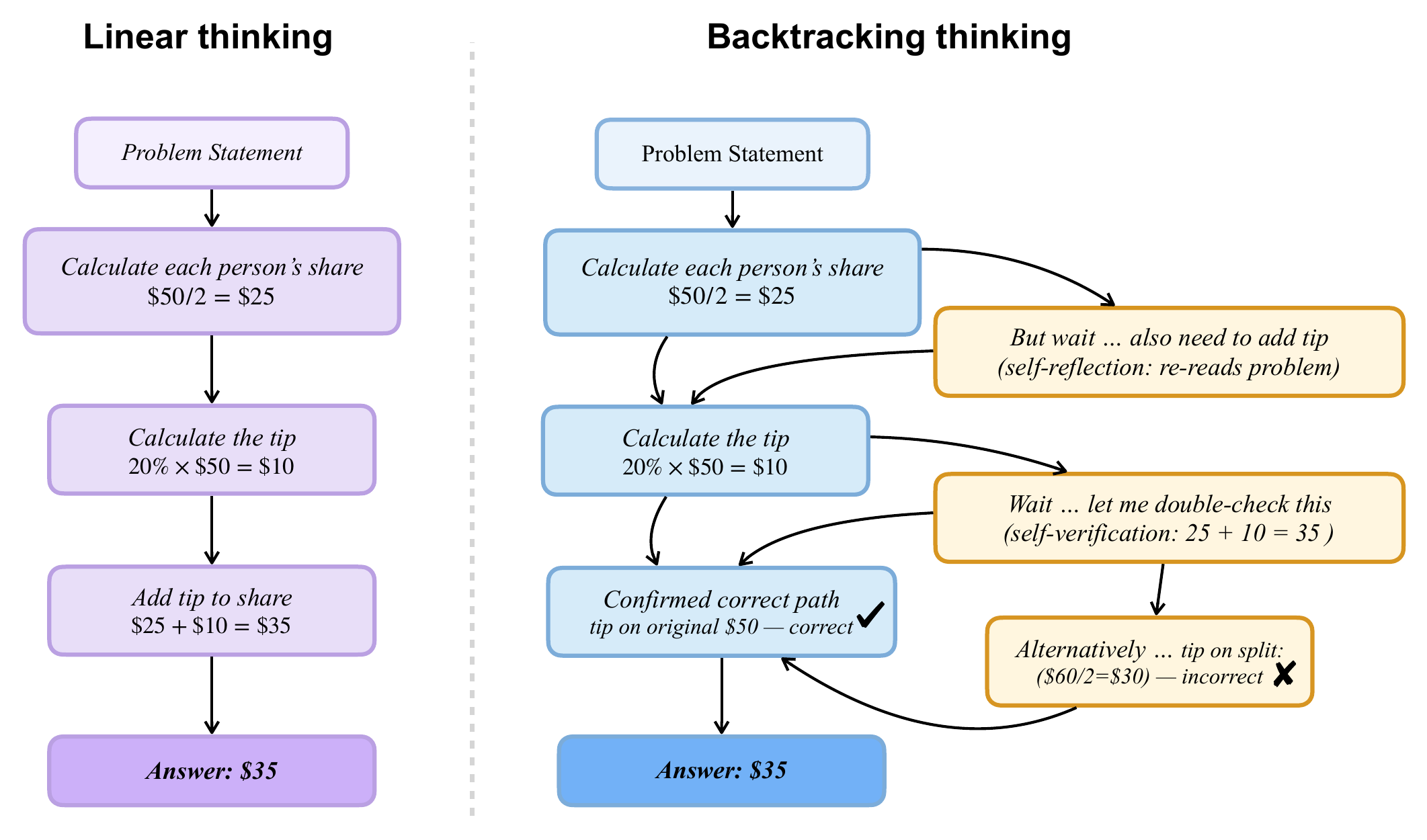}
\vspace{-1.0em}
\caption{
\looseness=-1 \small Illustrative examples of linear vs. backtracking reasoning modes.
}
\label{fig:reason-linear-backtrack}
\end{figure}

\looseness=-1 Another important mode of variation in reasoning is the \emph{structure} of the reasoning process.
We focus on two common modes: \textcolor{purple70}{\emph{linear thinking}} and \textcolor{cyan60}{\emph{backtracking thinking}} (Figure \ref{fig:reason-linear-backtrack}).  
In linear thinking, the model follows a direct, forward chain of steps without verification or revisiting earlier decisions (\textbf{left}). In contrast, backtracking thinking involves self-reflection and verification, where the model revisits intermediate steps (e.g., ``wait'',  ``let me check'' or ``alternatively,'') before committing to a solution (\textbf{right}). Figure~\ref{fig:reason-linear-backtrack} illustrates representative examples of these two reasoning modes.

\vspace{-0.8em}
\paragraph{Setup.} 
\looseness=-1 Rather than inducing this behavior with synthetic fine-tuning data, we study off-the-shelf R1-distilled models, including DeepSeek-R1-Distill-Qwen-1.5B and DeepSeek-R1-Distill-Llama-8B (additional Qwen-7B results in Appendix Section~\ref{sec_app:additional_results_distilled_reasoning_prefix_performance}). These models are trained on mixtures of DeepSeek-V3 and DeepSeek-R1 data, which exhibit linear and backtracking reasoning modes, respectively~\citep{guo2025r1}. This resembles our setup in Section~\ref{sec:exp-reason-lang}, suggesting that these models may similarly struggle to select the appropriate reasoning mode.

\begin{figure}[t]
\vspace{-2.0em}
\centering
\includegraphics[width=\linewidth]{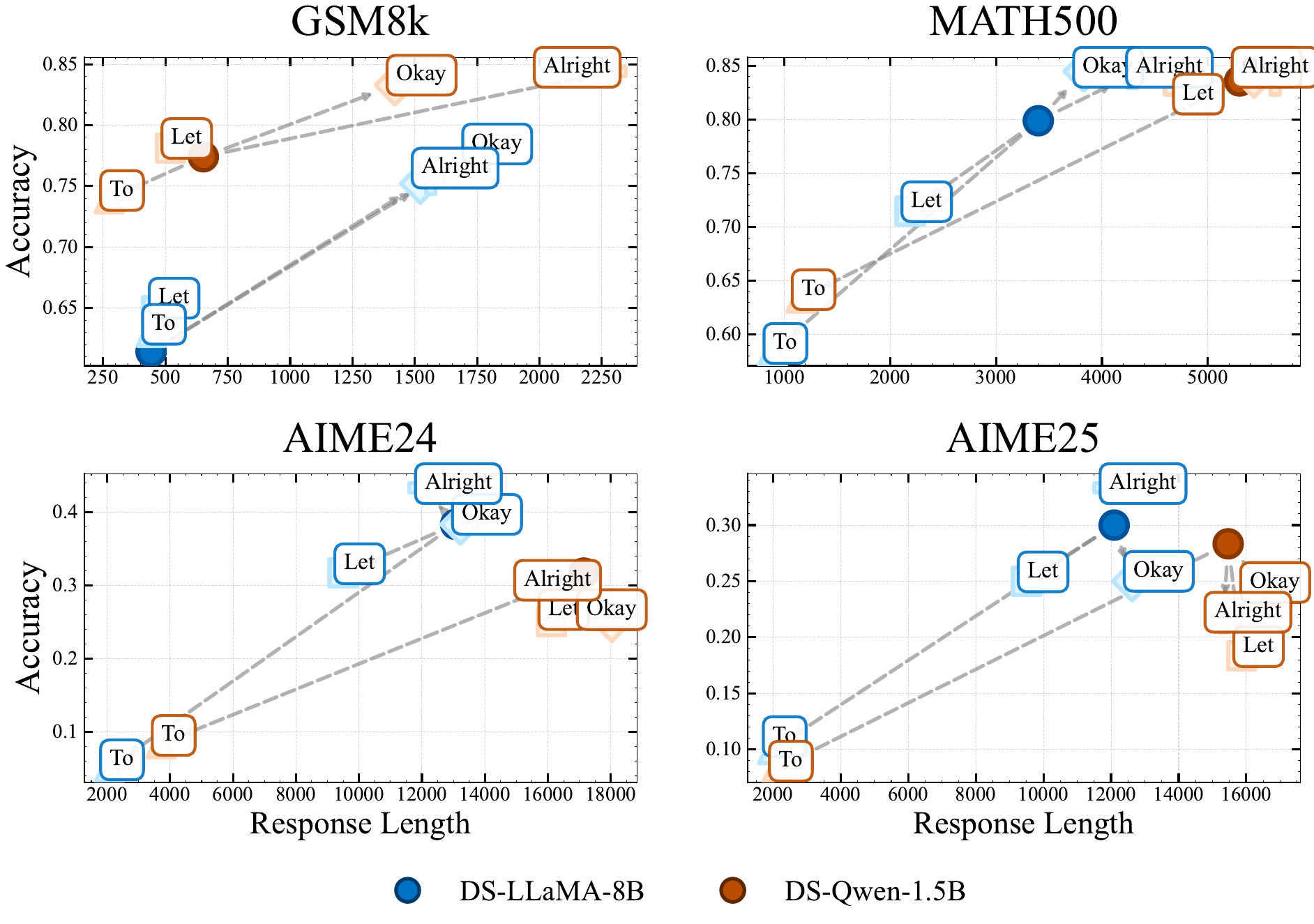}
\caption{
\looseness=-1 \small \textbf{Impact of prefix token manipulation on reasoning behavior.} 
Small changes only in one of the initial tokens (e.g., “Okay,” “Let,” “To”) lead to considerably large variations in response length and accuracy across different benchmarks datasets (GSM8K, Math500, AIME 2024/2025).
}
\label{fig:performance_length_bolder_multi_prefix}
\vspace{-1.5em}
\end{figure}

\paragraph{Results.} 
\looseness=-1 
By using CoT-without-prompting decoding mechanism~\citep{wang2024chainofthought}, we observe an interesting behavior where these models are \textbf{highly brittle at the \emph{initial tokens} of the reasoning trace}. Small variations in the first token (e.g., ``Okay,'' vs. ``Let'' or ``To'') can lead to drastically different reasoning behaviors out of the models, including changes in structure, length, and accuracy—even for the same input questions.
We interpret these \textbf{initial tokens as the practical realization of decision points}: they implicitly determine which reasoning mode the model commits to. Similar to indecipherable nodes in the graph branching experiments, the model here also appears to rely on weak or spurious cues (e.g., stylistic prefixes) to select a reasoning path at initial tokens as decision points.
In our experiments, we study this through systematically varying prefix tokens and evaluating their performance across multiple benchmark datasets (GSM8K, MATH-500, AIME24, and AIME25). The results in Figure~\ref{fig:performance_length_bolder_multi_prefix} show that small prefix changes can significantly alter both reasoning behavior and performance.
In particular, prefixes such as \emph{``Okay''} or \emph{``Alright''} consistently trigger backtracking behavior in thinking traces, leading to considerably longer responses with more verification steps and improved performance on math reasoning benchmarks. 

\looseness=-1 These results suggest that reasoning structure itself behaves like a fork in the road: the model must commit to either linear or backtracking thinking, and this choice largely determines both the trajectory and final outcome. 
We find that these decision points are brittle where small, seemingly irrelevant changes in the initial prefix tokens can shift the model between reasoning modes, leading to large differences in behavior and performance. 
This indicates that reasoning mode selection is driven by weak signals, making it a key source of instability and side effects such as coverage collapse, but also a natural point of intervention for control. 
In the next section (Sec~\ref{sec:mitigation}), we leverage this observation and show how manipulating prefix tokens can actually mitigate the coverage shrinkage by encouraging diversity over the reasoning modes.

\paragraph{Over-thinking and Under-thinking.} 
\looseness=-1 We also perform a fine-grained analysis showing that strategy selection in distilled reasoning models correlates with irrelevant lexical features rather than the problem structure. We evaluate this on (1)~simple knowledge questions (e.g., ``What is the capital of France?''), where the complex reasoning process is usually unnecessary, as well as (2)~counterfactual arithmetic questions (e.g., ``What is 6 + 3 in base-7?''), which require careful reasoning and verification.

\begin{wrapfigure}{r}{0.55\textwidth} 
\centering
\includegraphics[width=\linewidth]{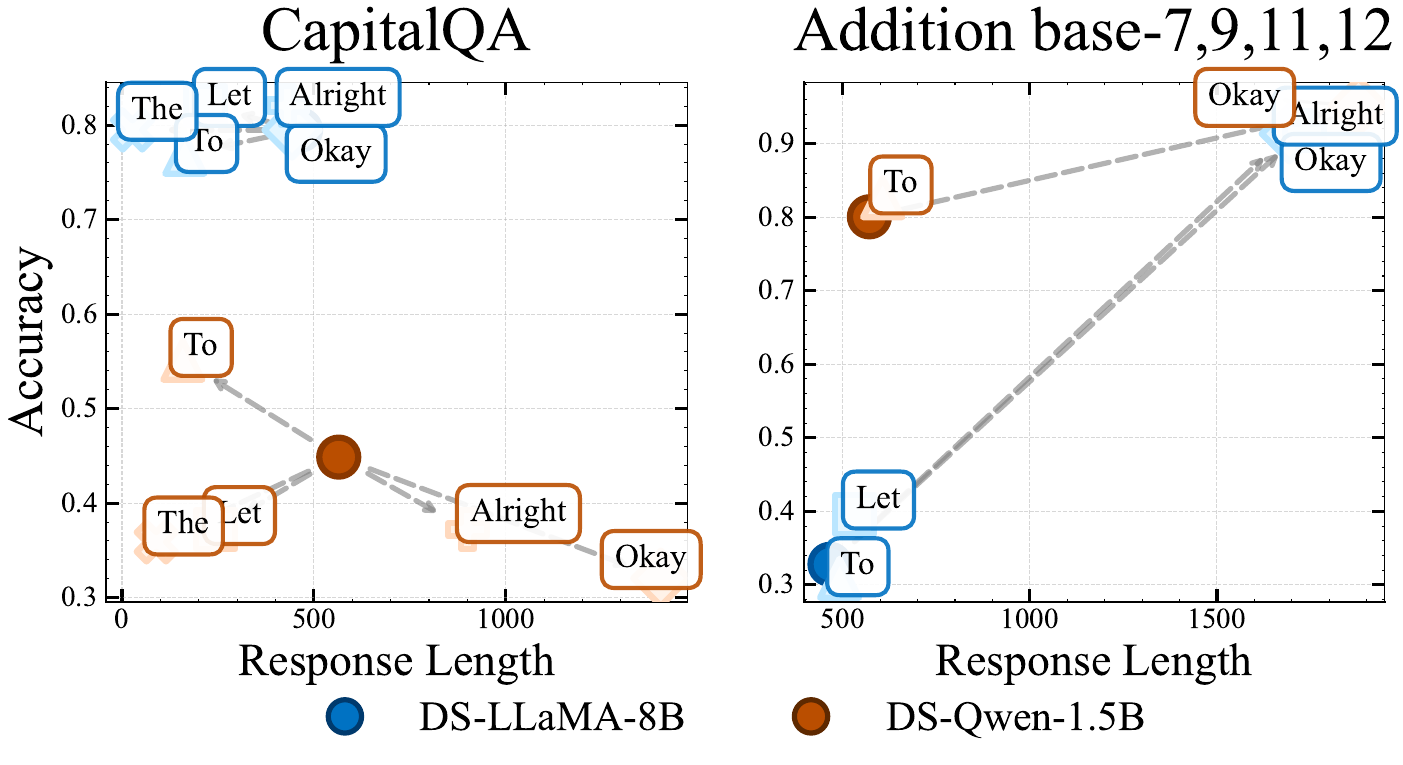}
\caption{
\small \looseness=-1 
Models' default behavior shows over-thinking on factual QA (backtracking hurts) and under-thinking on counterfactual arithmetic (backtracking helps).
}
\label{fig:performance_length_ovt_udt}
\end{wrapfigure}
An efficient approach is to perform linear, straightforward thinking for the first type of questions and backtracking thinking for the second type of questions.
However, our results (presented in Figure \ref{fig:performance_length_ovt_udt}) show that the models struggle to choose suitable thinking modes in these tasks.
Specifically, on counterfactual arithmetic problems, the model’s default behavior mostly resembles linear thinking, but enforcing non-linear or backtracking strategies (e.g., through prefixes like ``Okay'') can boost performance by up to 60\%. This shows clear \emph{under-thinking} in models when more reasoning and verification is actually needed. 
On the other hand, on CapitalQA factual problems, the model performs better with shorter and linear reasoning structure (e.g., through prefixes like “To”), yet its default behavior mostly resembles longer reasoning that hurts performance.  This shows \emph{over-thinking} in models when less reasoning is actually sufficient. More detailed examples regarding this observation is provided in Appendix Section \ref{sec:distilled_models_responses}.

\section{Discussion on Data-inspired Shrinkage Mitigation Strategies}
\label{sec:mitigation}

\looseness=-1 Our analysis demonstrates that coverage shrinkage is a data-driven artifact, where models learn to lock alternative reasoning paths behind spurious decision points. Building on this understanding, we explore simple strategies to mitigate this collapse by directly targeting these decision points.

\vspace{-0.8em}
\paragraph{Data Diversity Design.}
\looseness=-1 Our experiments in Section \ref{sec:exp-reason-lang} show that per-problem coverage of alternative decisions can significantly control shrinkage behavior compared to distributing the same level of diversity across the problems. This highlights the importance of post-training data design and distribution of diversity that explicitly accounts for reasoning decision-point structure and its coverage among problems. Our findings also explain why collecting multiple response modes per prompt is effective for training unified reasoning models~\citep{NemotronCascade_Scaling_Cascaded_Reinforcement_Learning}.

\vspace{-0.8em}
\paragraph{First-token Manipulation.}
\looseness=-1 Motivated by the observed brittleness of reasoning models to initial prefix tokens, we hypothesize that the \textbf{alternative reasoning strategies are still present inside the post-trained models, simply locked behind collapsed decision points}. 
To test this, we propose a simple inference-time intervention: instead of relying on the default generation, we perturb the initial decision by forcing the model to start from different high-probability initial reasoning tokens. Concretely, for each problem, we uniformly sample from the Top-$k$ candidate prefix tokens (Top-8 in our experiments) and continue solutions. We compare this to standard decoding (Default) and a deterministic Top-1 prefix setting.

\begin{figure*}[ht]
\centering
\includegraphics[width=\textwidth]{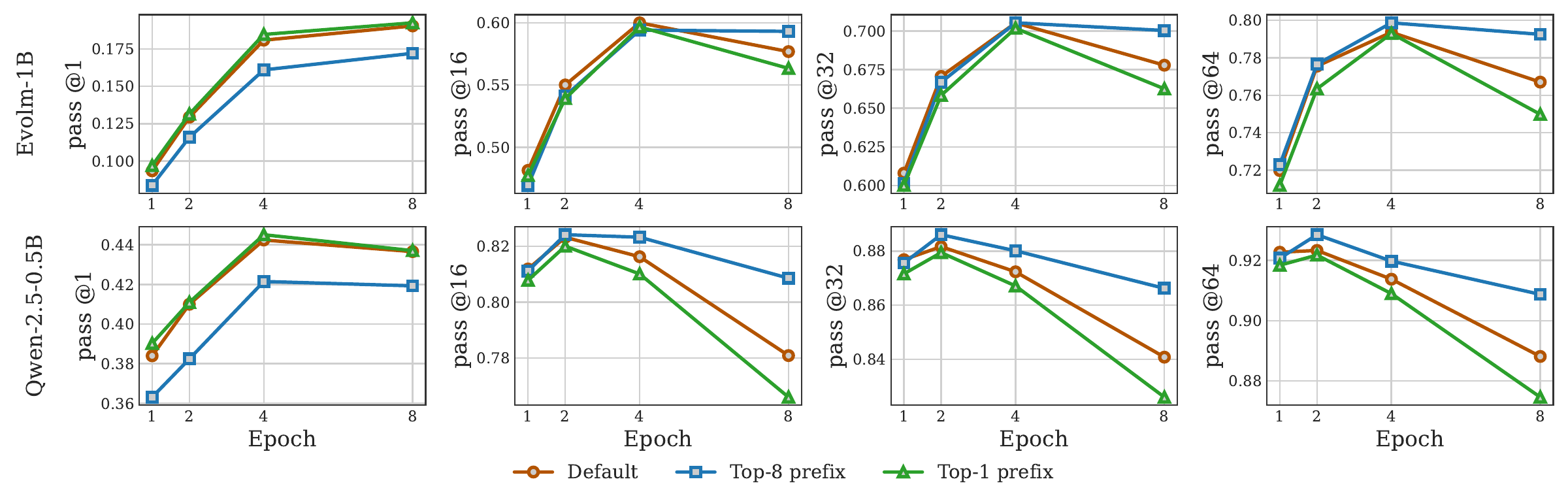}
\caption{
\small \textbf{Recovering coverage via prefix perturbation.} Top-$k$ prefix sampling (Top-8) mitigates coverage shrinkage and improves pass@k at larger $k$}
\label{fig:passk-first-token-gsm8k}
\end{figure*}

Figure~\ref{fig:passk-first-token-gsm8k} reports pass@k on GSM8K across SFT checkpoints for \texttt{EvoLM-1B} and \texttt{Qwen-2.5-0.5B} backbone models finetuned on a MetaMathQA subset \citep{yu2023metamath}. As observed, all pass@k improve during the early phase of training. However, in later stages, only pass@1 continues to improve while pass@k consistently declines across settings.
Our intervention shows that the Top-$k$ prefix setting (Top-8) matches early performance but achieves substantially higher pass@k later, effectively recovering lost coverage in post-trained models.
To distinguish this effect from generic increases in sampling entropy, we additionally compare against high-temperature decoding in Appendix~\ref{app:prefix-vs-temperature}. We find that targeted prefix perturbation consistently outperforms temperature scaling across both model families, particularly for moderate sampling budgets (pass@8--32), indicating that the gains arise from perturbing the model's initial routing decisions rather than simply increasing output diversity.

\vspace{-1.0em}
\section{Conclusion}
\vspace{-1.0em}

\looseness=-1 In this work, we investigate the phenomenon of ``coverage shrinkage'' in post-training reasoning language models. 
Diverging from prevailing theories that blame post-training algorithms, we propose a data-centric hypothesis and argue that shrinkage is highly driven by ``forks in the road'' or decision points in fine-tuning data where multiple valid reasoning paths exist, but the rationale for selecting one over the other is obscured.
Through experiments on controlled case studies, synthetic tasks, and mathematical reasoning, we demonstrate that models learn to rely on spurious features to resolve the ambiguity in reasoning chains, leading to coverage collapse after the post-training. By mitigating this shrinkage via targeted data diversity and inference-time prefix manipulations, we demonstrate that coverage and reasoning diversity is not lost after post-training, but merely locked behind the uncalibrated decisions. We hope that our insights provide a useful lens for those working on post-training for reasoning, data curation, or test-time scaling.

\section*{Acknowledgments}
The project was supported in part by Coefficient Giving (formerly Open Philanthropy). 
Preliminary work was conducted at the Center for AI Research at VinUniversity.

\bibliography{colm2026_conference}
\bibliographystyle{colm2026_conference}

\appendix
\section{Details on Synthetic Experiment}
\label{sec:app-graph}
\subsection{Task design}
We introduce a synthetic task, designed to capture two core components of mathematical problem solving: (1) chains of variable manipulation and (2) planning over a dependency structure. In this task, the prompt specifies a set of functional dependencies among variables together with a query about the value of a designated target variable. During the SFT phase, the model is trained on solutions consisting of a sequence of intermediate computation steps in order to find the target's value.
\begin{minipage}[t]{0.68\linewidth}
\vspace*{0pt}
\small
\textbf{Q:} Let each letter represent a numerical variable. These variables are defined as follows:
$n=10$; $m=n+12$; $k=m+3$; $h=k + 4$; $l=n+19$; $j=l+17$; $x=j+2$.
What is the resulting value of $x$?
\end{minipage}\hfill
\begin{minipage}[t]{0.28\linewidth}
\vspace*{0pt}
\centering
\begin{tikzpicture}[
  scale=0.55,
  every node/.style={
    circle,
    draw,
    minimum size=6mm,
    inner sep=0pt,
    font=\scriptsize
  },
  edge from parent/.style={draw, -Stealth},
  level 1/.style={sibling distance=2.2cm},
  level 2/.style={sibling distance=1.6cm}
]
\node[draw=green, thick] (n) {n}
  child {
    node (m) {m}
    child { node (k) {k} 
        child { node (h) {h}}
    }
  }
  child {
    node (l) {l}
    child {
      node (j) {j}
      child { node[draw=red, thick] (x) {x} }
    }
  };
\end{tikzpicture}
\end{minipage}

A key property of this task is the topology of the underlying dependency graph. In the example shown above, we consider a \emph{path-star} graph, introduced in \citet{bachmann2024the} to study the failure of the teacher-forcing learning objective. The graph consists of a central root node with multiple outgoing paths. The target variable lies at the end of a unique path. To determine its value, the model must first identify the relevant dependency path and then correctly compute all intermediate variables along that path.

\paragraph{Prompts.}
\label{sec:app-graph-prompt}
We use the Alpaca instruction format. An example of a prompt is shown below:

\begin{tcolorbox}[colback=gray!25,colframe=gray!75,title=A problem from the synthetic task]
Below is an instruction that describes a task, paired with an input that provides further context. Write a response that appropriately completes the request.
\\
\\
\textbf{Instruction:} \\
Solve the following math problem, and put your final answer within \verb|\boxed{}|.
\\
\\
\textbf{Input:} \\
Consider a system of variables where each variable is defined as follows:
t = x + 10, e = c + 17, s = y + 13, x = a + 2, b = v + 16, j = s + 17, r = d + 18, u = o + 20, h = f + 6, o = v + 6, i = p + 16, y = u + 11, a = h + 5, v = 3, d = z + 13, z = e + 13, g = i + 12, f = m + 11, c = j + 6, p = t + 6, m = b + 3.
If $v = 3$, determine the value of $g$.
\\
\\
\textbf{Response:}\\
\end{tcolorbox}

We also provide examples of three variations of the solution in the following.
\label{sec:app-graph-type}
\\
\\
\begin{tcolorbox}[colback=gray!25,colframe=gray!75,title=Example of \textit{Forward} Solution (with decision point)]
To find the target value, we compute the following variables step by step:
\begin{enumerate}
    \item $o = d + 19 = 37$
    \item $k = o + 9 = 46$
    \item $w = k + 1 = 47$
    \item $v = w + 6 = 53$
    \item $g = v + 14 = 67$
    \item $f = g + 11 = 78$
    \item $m = f + 9 = 87$
    \item $l = m + 5 = 92$
    \item $p = l + 7 = 99$
    \item $s = p + 11 = 110$
\end{enumerate}
Thus, $s = \boxed{110}$.
\end{tcolorbox}

\begin{tcolorbox}[colback=gray!25,colframe=gray!75,title=Example of \textit{Reverse} Solution (without decision point)]
To find the target value, we compute the following variables step by step:
\begin{enumerate}
    \item Substitute $p = l + 7$ into the target expression, yielding $s = l + 18$.
    \item Substitute $l = m + 5$ into the target expression, yielding $s = m + 23$.
    \item Substitute $m = f + 9$ into the target expression, yielding $s = f + 32$.
    \item Substitute $f = g + 11$ into the target expression, yielding $s = g + 43$.
    \item Substitute $g = v + 14$ into the target expression, yielding $s = v + 57$.
    \item Substitute $v = w + 6$ into the target expression, yielding $s = w + 63$.
    \item Substitute $w = k + 1$ into the target expression, yielding $s = k + 64$.
    \item Substitute $k = o + 9$ into the target expression, yielding $s = o + 73$.
    \item Substitute $o = d + 19$ into the target expression, yielding $s = d + 92$.
    \item Substitute $d = 18$ into the target expression, yielding $s = 110$.
\end{enumerate}
Thus, $s = \boxed{110}$.
\end{tcolorbox}

\subsection{Additional control setting}

To control for possible confounding factors introduced by reverse traversal, we construct an additional synthetic variant that preserves the original forward computation order while explicitly annotating each branching decision with a short rationale indicating why the correct branch is selected. This intervention removes the ambiguity at decision points without altering the dependency graph or reasoning direction.

\begin{tcolorbox}[colback=gray!25,colframe=gray!75,title=Example of \textit{Forward} Solution with an explicit path-selection rationale]
\verb|<rationale>|\\
Since $s$ depends on $p,l,m,f,g,v,w,k$ and $o$, I choose branch $o$.\\
\verb|</rationale>| \\
To find the target value, we compute the following variables step by step:
\begin{enumerate}
    \item $o = d + 19 = 37$
    \item $k = o + 9 = 46$
    \item $w = k + 1 = 47$
    \item $v = w + 6 = 53$
    \item $g = v + 14 = 67$
    \item $f = g + 11 = 78$
    \item $m = f + 9 = 87$
    \item $l = m + 5 = 92$
    \item $p = l + 7 = 99$
    \item $s = p + 11 = 110$
\end{enumerate}
Thus, $s = \boxed{110}$.
\end{tcolorbox}

Table~\ref{tab:graph-rationale} shows that this simple modification completely changes the optimization dynamics. Unlike the original forward setting, pass@$k$ remains near perfect throughout post-training for both Qwen2.5-0.5B and EvoLM-1B, matching the behavior of the Reverse (w/o DP) setting. These results provide stronger evidence that the observed coverage collapse originates from implicit decision points in the supervision rather than changes in reasoning direction or task complexity.

\begin{table}[t]
\centering
\small
\caption{\textbf{Forward reasoning with explicit decision rationale.}
Providing an explicit rationale at each decision point prevents coverage collapse while preserving the original forward computation order.}
\label{tab:graph-rationale}
\begin{tabular}{llcccc}
\toprule
Model & Epoch & Pass@1 & Pass@2 & Pass@8 & Pass@32\\
\midrule
\multirow{5}{*}{\texttt{Qwen2.5-0.5B}}
&1&0.74&0.87&0.96&0.99\\
&2&0.88&0.98&1.00&1.00\\
&4&0.93&0.99&1.00&1.00\\
&8&0.93&0.99&1.00&1.00\\
&16&0.93&0.99&1.00&1.00\\
\midrule
\multirow{5}{*}{\texttt{EvoLM-1B}}
&1&0.32&0.48&0.79&0.94\\
&2&0.65&0.82&0.96&0.99\\
&4&0.75&0.91&0.99&1.00\\
&8&0.79&0.94&0.99&1.00\\
&16&0.80&0.94&1.00&1.00\\
\bottomrule
\end{tabular}
\end{table}

\subsection{Experimental settings}
\paragraph{Dataset.} We generate a list of variables and equations from a star graph with 2 branches, and the lengths from the root to leaf nodes are 10. Next, we choose text templates and generate a pair of questions and ground truth solutions. This process results in a training set of 6400 samples and a testset of 1000 samples.

\paragraph{Training and Evaluation.} We supervise finetune the pretrained Qwen-2.5-0.5B model \citep{qwen2.5} and EvoLM-1B \citep{qi2025evolm} on the training set for 16 epochs with lr=2e-5. For evaluation, we mark a solution as correct if the boxed output matches the value of the target variable.

\subsection{Probing confidence at decision points}
\label{sec:app-graph-conf}

We probe the confidence of the model at the decision points of the forward solution by computing the probability of the next token of the following prompt,

\begin{tcolorbox}[colback=gray!25,colframe=gray!75]
Below is an instruction that describes a task, paired with an input that provides further context. Write a response that appropriately completes the request.
\\
\\
\textbf{Instruction:} \\
Solve the following math problem, and put your final answer within \verb|\boxed{}|.
\\
\\
\textbf{Input:} \\
Consider a system of variables where each variable is defined as follows:
t = x + 10, e = c + 17, s = y + 13, x = a + 2, b = v + 16, j = s + 17, r = d + 18, u = o + 20, h = f + 6, o = v + 6, i = p + 16, y = u + 11, a = h + 5, v = 3, d = z + 13, z = e + 13, g = i + 12, f = m + 11, c = j + 6, p = t + 6, m = b + 3.
If $v = 3$, determine the value of $g$.
\\
\\
\textbf{Response:}\\
To find the target value, we compute the following variables step by step:
\begin{enumerate}
    \item 
\end{enumerate}
\end{tcolorbox}

Since the model is trained to predict the name of the variable of the first step after this input, we are able to identify which branch the model will take subsequently and the confidence of following this branch.

\subsection{Learning Spurious Cues for Branch Selection.}
\label{sec:appendix_spurious_cues_graph}
\begin{wrapfigure}{r}{0.4\textwidth} 
\centering
\vspace{-0.5em}
\includegraphics[width=1.0\linewidth]{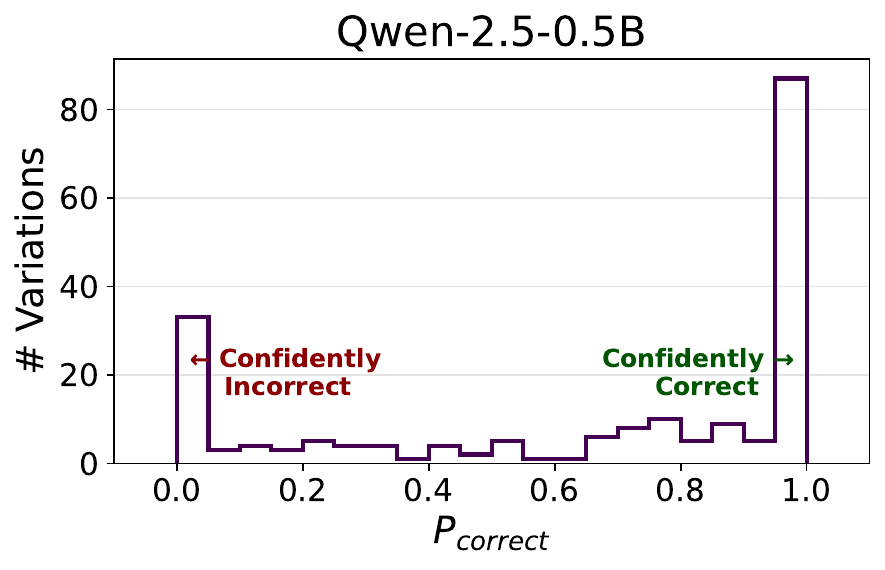}
\vspace{-2.0em}
\caption{
\looseness=-1 \small Model confidence at decision points across prompt variations with identical semantics per problem.
}
\label{fig:graph_shuffle_rules_confidence}
\end{wrapfigure}

\looseness=-1 Previous work shows that reasoning in language models is highly sensitive to minor input changes \citep{mirzadeh2025gsmsymbolic, jiang2024peek}. We further examine this by probing the model’s branch selection: we perturb prompts by shuffling variable dependencies (e.g., \texttt{n=10; m=n+12; k=m+3} to \texttt{m=n+12; n=10; k=m+3}), preserving semantics but altering surface formats. We observe that despite the same logic, these changes can significantly shift the model’s behavior, causing it to select different reasoning branches (as shown in Figure~\ref{fig:graph_shuffle_rules_confidence}). This suggests the model may rely on spurious cues like dependency order when it encounters decision points. We provide more details regarding this fragility in distilled reasoning models and their thinking structure in Section \ref{sec:exp-reason-structure}.

\subsection{Reinforcement Learning with Verification Reward (RLVR)}
\label{sec:grpo_graph}

\begin{figure}
    \centering
    \includegraphics[width=0.9\linewidth]{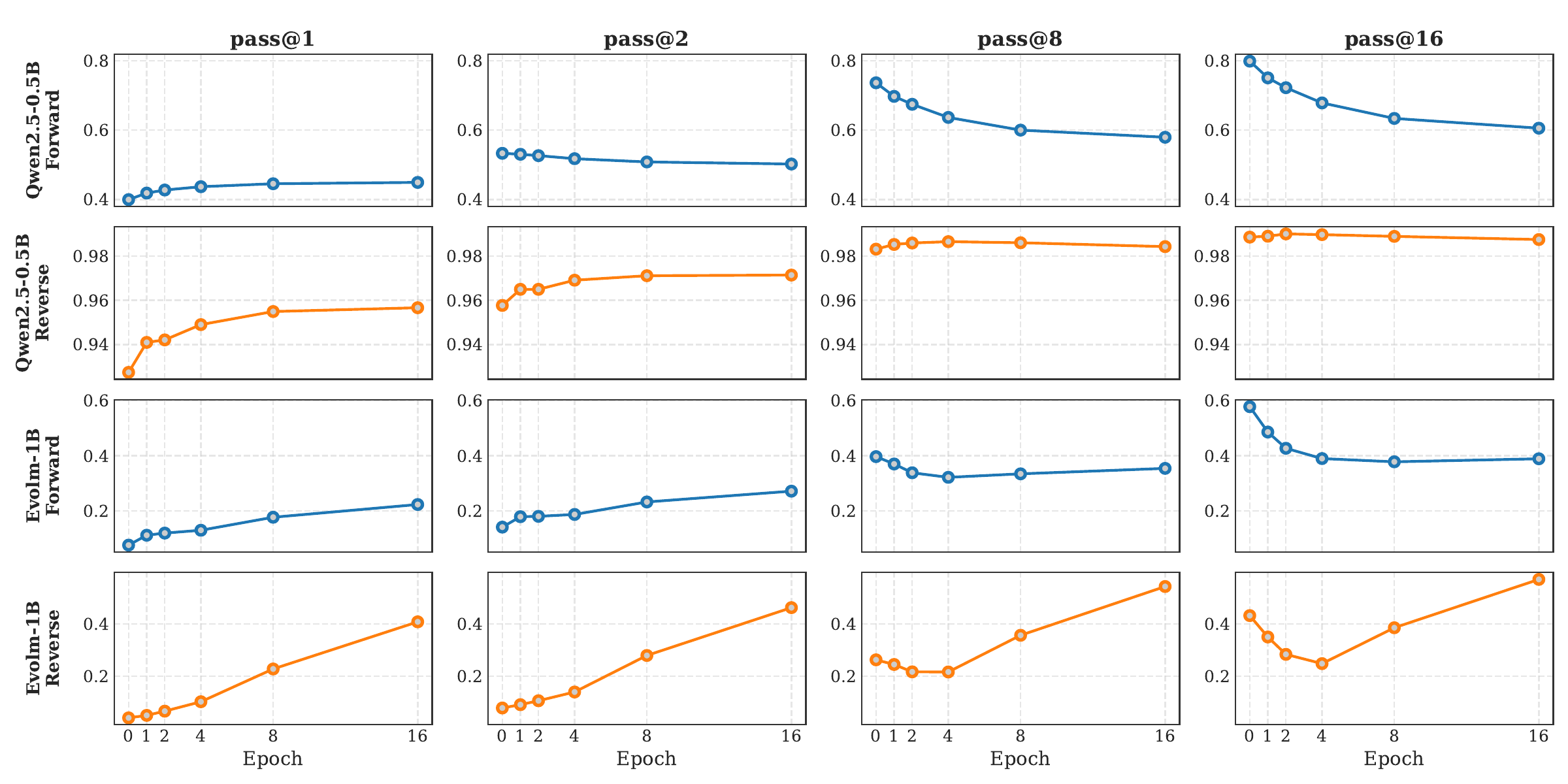}
    \caption{Pass@k performance when running GRPO on models pretrained on forward and reverse (-DP) solutions.}
    \label{fig:grpo_passk_synthetic}
\end{figure}

Our experiments in the main text focus on SFT dynamics and pass@k across SFT epochs. To examine the role of data and indecipherable decision points under other post-training methods, we apply a standard RLVR algorithm, Group Relative Policy Optimization (GRPO) \citep{shao2024deepseekmath}, on models that are first supervised fine-tuned for one epoch on either forward (with decision points) or reverse (without decision points) data. The results, shown in Figure \ref{fig:grpo_passk_synthetic}, demonstrate that the same coverage shrinkage pattern emerges during the RL phase when the training data contains decision points (the Forward setting). These findings shed insight into how pretraining data affects coverage shrinkage during RL; it arises when the optimization is performed over responses that require navigating ambiguous decision points. In contrast, when such blind forks are absent, the RL algorithm preserves coverage.

\section{Details on Experiment with Reasoning Modes}

\subsection{Natural Language vs. Code Reasoning}
In this experiment, we collect SFT samples from two common mathematical reasoning datasets:  
OpenMathInstruct-1 \citep{toshniwal2024openmath}, and OpenMathInstruct-2 \citep{toshniwal2024openmath2}. 
OpenMathInstruct-2 consist of solutions in natural language, while OpenMathInstruct-1 consists of Python code solutions generated by Mixtral-8x7B \citep{jiang2024mixtral}. 
We first filter only GSM8k questions and combine them from these datasets.
We sample the solution such that each reasoning style (NL vs Code) is distributed equally in the whole dataset.
Questions in the test set from GSM8k are used for evaluation.

For the sampling implementation and parameters, we use the default vllm inference server \citep{kwon2023efficient} with $temperature=1.0,\;top\_p = 0.95,\;max\_new\_tokens=1024$ and generate 64 solutions per test sample.
The backbone models we used in this experiment are: \texttt{Qwen-2.5-0.5b}~\citep{qwen2.5}, \texttt{EvoLM-1B-160BP-8M42M} and \texttt{EvoLM-4B-160BP-8M42M}~\citep{qi2025evolm}.

\subsection{Additional experimental results on reasoning modes}

In this section, we provide additional experimental results when fine-tuning an Olmo-7B base model, allowing us to evaluate our hypotheses on a different model family.

\begin{figure}
    \centering
    \includegraphics[width=0.95\linewidth]{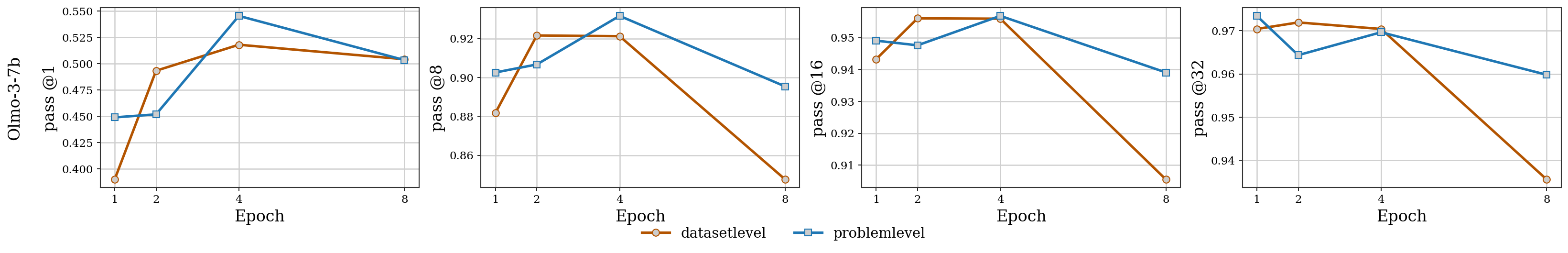}
    \caption{
\small \textbf{Additional results on efffect of data diversity design on coverage in reasoning mode selection task}. Pass@k evaluated on the GSM8K test split across SFT epochs for \textcolor{rose60}{\emph{data-level}} vs. \textcolor{cyan70}{\emph{problem-level}} data diversity.}
    \label{fig:passk_epochs_olmo-3-7b}
\end{figure}

The results (shown in Figure \ref{fig:fraction_code_olmo7b_2modes} and Figure \ref{fig:passk_epochs_olmo-3-7b}) are consistent with our central claims. Under dataset-level diversity, Olmo-7B exhibits the same mode-collapse behavior observed in our original experiments: despite exposure to diverse reasoning traces at the dataset level, the model increasingly concentrates probability mass on a small subset of reasoning modes during training. In contrast, when diversity is at the problem level, coverage is preserved and the collapse is substantially mitigated.

\begin{figure}
    \centering
    \includegraphics[width=0.85\linewidth]{assets//styles/fraction_code_olmo7b_2modes.pdf}
    \caption{\looseness=-1 \small \textbf{Additional results on reasoning mode preference under different diversity designs.}
Distribution of per problem code vs. NL reasoning across SFT epochs  for \textcolor{rose60}{\emph{data-level}} vs. \textcolor{cyan70}{\emph{problem-level}} diversity.}
    \label{fig:fraction_code_olmo7b_2modes}
\end{figure}

\subsection{Distilled Reasoning Models}
\label{sec_app:additional_results_distilled_reasoning_prefix_performance}

\paragraph{Decoding Configuration.}
In this experiment, we use the \texttt{vLLM} library \citep{kwon2023efficient} for efficient inference with the following parameters:
\begin{itemize}
    \item \textbf{Temperature:} 0.6
    \item \textbf{Top-$p$:} 0.95
    \item \textbf{Top-$k$:} -1
    \item \textbf{Max tokens:} 32768
    \item \textbf{Max model length:} 32768
    \item \textbf{Trust remote code:} \texttt{True}
\end{itemize},
which are recommended by DeepSeek.
To format the input prompt, we use each model's default chat template.

\begin{figure}
\centering
\includegraphics[width=0.95\linewidth]{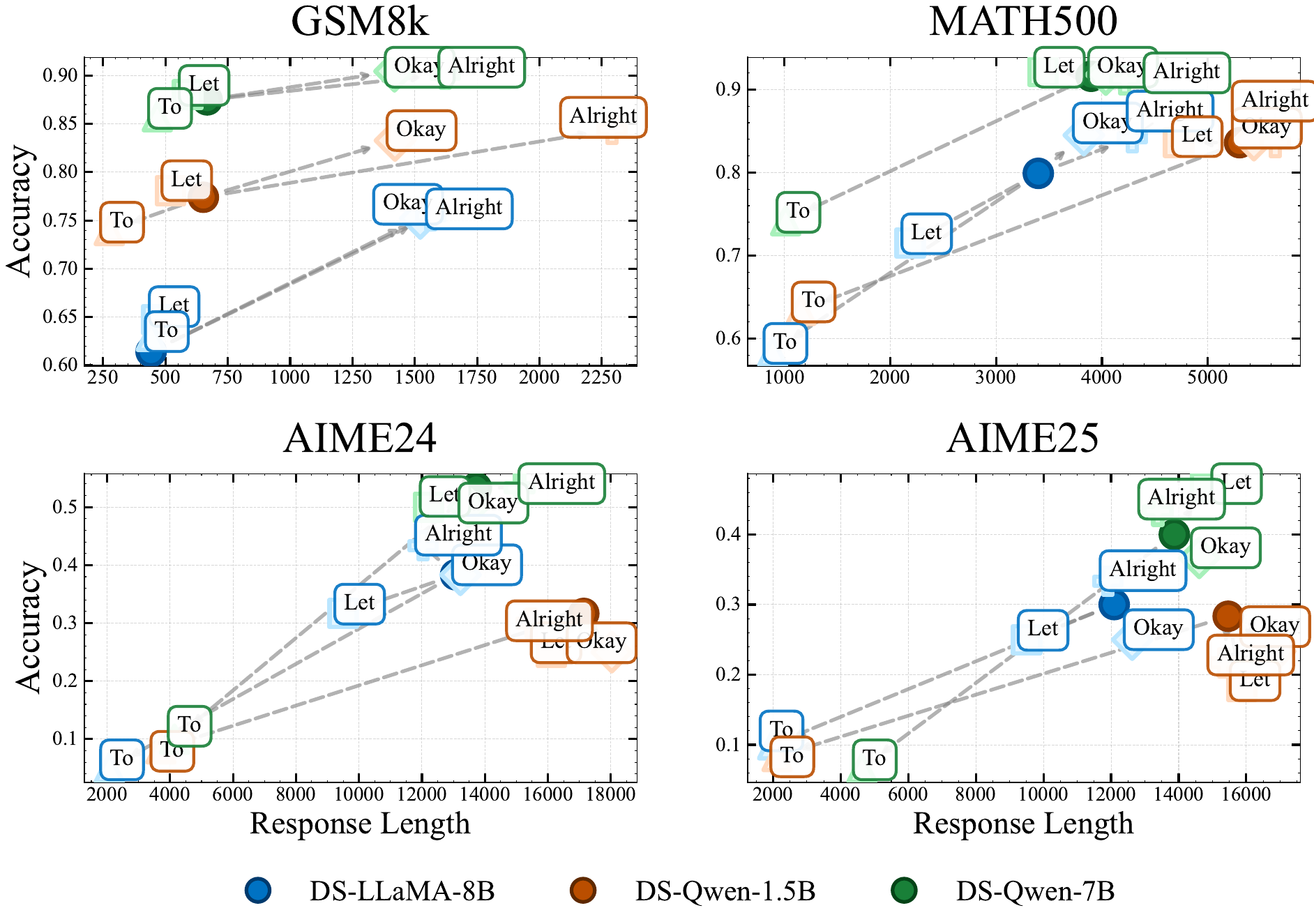}
\vspace{-0.5em}
\caption{\textbf{Performance and Response Length across different prompt prefixes.} The accuracy of two models—DS-Qwen-1.5B, DS-Qwen-7B, and DS-LLaMA-8B—across mathematical benchmarks (GSM8K, MATH500, AIME24, and AIME25). The data demonstrates a high degree of performance variance (up to 20\%) and response length variance (up to 5.8x) depending entirely on the starting word or phrase of the thinking prefix.}
\label{app_fig:performance_length_bolder_multi_prefix_extended}
\vspace{-1.0em}
\end{figure}

Figure \ref{app_fig:performance_length_bolder_multi_prefix_extended} presents the performance of DeepSeek-R1-Distilled reasoning models when generating responses from different thinking prefixes across 4 different reasoning benchmark GSM8k \citep{cobbe2021gsm8k}, MATH500 \citep{lightman2023lets}, AIME24 \citep{aime24}, and AIME25 \citep{aime25}.

\subsection{Examples of models' responses}
\label{sec:distilled_models_responses}

For the counterfactual addition prompt, we use the template from \citet{wu2024reasoningreciting}. In the following, we provide examples of questions and responses from models:

\begin{tcolorbox}[colback=white, colframe=black, sharp corners, boxrule=0.5pt, title=Addition in base-12]
\textbf{Question:} You are a mathematician. Assuming that all numbers are in base-7 where the digits are "0123456", what is 66 + 50? \\

Let’s think step by step, and end the response with the result in \textbackslash boxed\{\}.\\

\textbf{Correct answer:} 146 
\end{tcolorbox}
\begin{lstlisting}[caption={Response from DS-LLaMA-8B}]
<think>
First, I need to add the two base-7 numbers 66 and 50.

I'll start by aligning the numbers vertically, ensuring each digit is in the correct place value.

Next, I'll add the digits in each column, starting from the rightmost digit.

In the units place, 6 plus 0 equals 6.

In the sevens place, 6 plus 5 equals 11. Since we're working in base 7, 11 divided by 7 is 1 with a remainder of 4. I'll write down the remainder and carry over the quotient.

Finally, I'll add the carried over 1 to the next higher place, resulting in a final sum of 116 in base 7.
</think>

**Solution:**

To add the two base-7 numbers **66** and **50** in base-7, follow these steps:

1. **Write the numbers vertically, aligning the digits by place value:**

```
  6 6
+ 5 0
--------
```

2. **Add the digits in each column, starting from the right (units place):**
   
   - **Units place:** 6 + 0 = 6
   
   - **Sevens place:** 6 + 5 = 11

3. **Since we're in base-7:**
   
   - 11 in base-10 is equal to **1** group of **7** (since \(11 \div 7 = 1\) with a remainder of 4).
   
   - Write down the remainder **4** and carry over the quotient **1** to the next higher place.

4. **Add the carried over value to the next higher place:**
   
   - The carried over **1** is added to the next (sevens) place: 1 (from the carry) + 1 (from the carry) = **2**

5. **Write down the final result:**
   
   Combining all the results, we get **116** in base-7.

**Final Answer:**

\[
\boxed{116}
\]
\end{lstlisting}

\begin{lstlisting}[caption={Response from DS-LLaMA-8B with "Okay" prefix}]
<think>
Okay, so I have this problem here: I need to add two numbers, 66 and 50, but both numbers are in base-7. Hmm, base-7, that's not our usual base-10 system we use every day. I remember that in base-7, each digit represents a power of 7, just like in base-10 each digit is a power of 10. So, to add these numbers, I guess I need to first convert them into base-10, add them, and then convert back to base-7. Or maybe I can add them directly in base-7 without converting. I'm not sure which is easier. Let me think.

First, let me recall how base-7 addition works. In base-7, each digit can only be from 0 to 6. So, if I add two digits and the sum is 7 or more, I have to carry over to the next higher place value. For example, if I'm adding 5 + 6 in base-7, that would be 11 in base-10, which is 1*7 + 4, so I would write down 4 and carry over 1. Okay, got it.

So, let's write down the numbers vertically, aligning the digits by their place values. Let me write 66 and 50 in base-7:

```
  6 6
+ 5 0
--------
```

Wait, hold on. Let me clarify the place values. In base-7, the rightmost digit is the units place (7^0), and the next is the sevens place (7^1), then 49s (7^2), and so on. So, 66 in base-7 is 6*7^1 + 6*7^0, which is 42 + 6 = 48 in base-10. Similarly, 50 in base-7 is 5*7^1 + 0*7^0, which is 35 + 0 = 35 in base-10. So, adding them together, 48 + 35 = 83 in base-10. Then, converting 83 back to base-7: 7^2 is 49, so 83 divided by 49 is 1 with a remainder of 34. 34 divided by 7 is 4 with a remainder of 6. So, 83 in base-7 is 1*49 + 4*7 + 6 = 1 4 6, so 146 in base-7.

Wait, but maybe I should try adding them directly in base-7 without converting to base-10. Let me try that. So, writing them vertically:

```
  6 6
+ 5 0
--------
```

Starting from the rightmost digit: 6 + 0. That's easy, 6. Then, moving to the left, 6 + 5. Hmm, 6 + 5 is 11 in base-10. But in base-7, 11 divided by 7 is 1 with a remainder of 4. So, I write down 4 and carry over 1. There are no more digits to add, so I just add the carried-over 1 in front. So, the result is 1 4 6, which is 146 in base-7. That matches my previous result. Okay, that's reassuring.

Wait, let me make sure I didn't make any mistakes. Let me check each step again. Starting from the right: 6 + 0 is 6, no carry-over. Then, 6 + 5 is 11 in base-10. Since we're in base-7, 11 divided by 7 is 1 with a remainder of 4. So, we write down 4 and carry over 1. Since there are no more digits, we just add the carried-over 1 at the beginning. So, yes, 146 in base-7. Converting back to base-10, 1*49 + 4*7 + 6*1 = 49 + 28 + 6 = 83. Which is correct because 66 in base-7 is 48 and 50 in base-7 is 35, 48 + 35 = 83. So, both methods give the same result.

I think I got it right. So, the answer is 146 in base-7, which is 83 in base-10. Therefore, the final answer is 146 base-7.
</think>

To solve the addition of 66 and 50 in base-7, we can perform the addition directly in base-7:

```
  6 6
+ 5 0
--------
```

1. Add the rightmost digits: 6 + 0 = 6 (no carry-over).
2. Add the next digits: 6 + 5 = 11 in base-10. Since we're in base-7, 11 divided by 7 is 1 with a remainder of 4. Write down 4 and carry over 1.
3. Add the carried-over 1 to the front.

Result: 1 4 6 in base-7.

So, the sum of 66 and 50 in base-7 is \boxed{146}.
\end{lstlisting}

\begin{tcolorbox}[colback=white, colframe=black, sharp corners, boxrule=0.5pt,title=Example of CapitalQA]
\textbf{Question:} What is the capital of Greece? \\
\textbf{Correct answer:} Athens 
\end{tcolorbox}

\begin{lstlisting}[caption={Response from DS-Qwen-1.5B with "The" prefix}]
<think>
  The capital of Greece is Athens. It is the most populous city in the country and the capital of both the Peloponnese and the island of Greece.
</think>

The capital of Greece is Athens.
\end{lstlisting}

\begin{lstlisting}[caption={Response from DS-Qwen-1.5B with "Okay" prefix}]
<think>
Okay, so I need to figure out the capital of Greece. Hmm, I remember that Greece is a country in Europe, but I'm not exactly sure where its capital is. I think it's a major city, maybe in a coastal area because of the sea. I've heard of places like Athens and maybe Thessaloniki, but I'm not certain. I should probably break this down step by step.

First, I'll start by recalling any information I know. I know that Greece is often referred to as the ""Venice of the Mediterranean."" I think that's a nickname, but I'm not sure if that helps. I also remember that Athens is the capital of Greece. Wait, but I think there's another city called Thessaloniki. Maybe that's another capital? Or perhaps I'm mixing up capitals with something else.

I should consider the geography. Greece has a lot of islands, including the Peloponnese, which runs from Sparta to Athens. So, Athens is a major city there. I think Thessaloniki is located in the southern part of the country, maybe in the Ionian Sea. But I'm not sure if it's the capital.

I think the capital is actually Thessaloniki. But I should verify this. I remember reading somewhere that the capital is called the ""City of the North"" because of its location in the north of Greece. But I'm not entirely sure. Maybe I can think of other cities in Greece and see which one is more likely to be the capital.

I know that Athens is the capital of Athens, which is a major city in Greece, and it's the capital of another country, but that's not relevant here. Maybe Thessaloniki is more of a regional capital or a city-state. I think in some contexts, especially in discussions about Greece, Thessaloniki is considered the capital, but I'm not certain.

I should also consider the population and the size of the city. I believe Thessaloniki has a significant population, and it's a major city in the south of Greece. Athens is a much larger city, but it's also a major city. So, which one is the actual capital?

I think I've heard conflicting information before. Sometimes, people refer to Athens as the capital, but others say it's Thessaloniki. I should try to find a reliable source or think about the usual way capitals are named. In many European countries, the capital is named after the capital of the country, but I'm not sure if that applies here.

Wait, I think in the case of Greece, the capital is actually Thessaloniki. It's the capital of the island of Thessalonica, which is part of Greece. So, maybe the capital is called Thessaloniki, but it's the capital of the island. That makes sense because the island is in the south of the country, so the capital would be in the south.

Alternatively, I could think about the layout of Greece. The capital is in the south, near the mainland, so it's more likely to be a regional capital. But I'm not entirely sure. I think the capital is actually Thessaloniki, so I should go with that.
</think>

The capital of Greece is Thessaloniki.
\end{lstlisting}

\section{Prefix Perturbation versus Temperature Sampling}
\label{app:prefix-vs-temperature}

One possible explanation for the gains from prefix perturbation is that it merely increases the diversity of generated outputs, similar to using a higher sampling temperature. To distinguish these hypotheses, we compare our method with standard temperature sampling.

Specifically, we evaluate two decoding temperatures ($T=1.3$ and $T=1.6$) while keeping all other decoding settings identical. We compare these baselines with our Top-8 prefix perturbation strategy across SFT checkpoints of both \texttt{Qwen-2.5-0.5B} and \texttt{EvoLM-1B}. Tables~\ref{tab:prefix-vs-temperature} reports the resulting pass@$k$ scores.

Overall, prefix perturbation consistently achieves higher pass@$k$ than temperature sampling across both model families. The advantage is most pronounced for practical sampling budgets (pass@8--32), where prefix perturbation improves coverage while largely preserving pass@1. Although higher temperatures occasionally match or slightly exceed prefix perturbation for pass@64 after convergence, they generally suffer a noticeable degradation in pass@1 and lower-budget pass@$k$.

These observations suggest that the effectiveness of prefix perturbation is not simply due to injecting additional randomness into decoding. Instead, perturbing the initial reasoning token more effectively explores alternative reasoning trajectories, supporting our hypothesis that post-training suppresses multiple reasoning strategies by collapsing early decision points rather than eliminating them entirely.

\begin{table*}[t]
\centering
\small
\caption{Comparison between Prefix Perturbation (Top-8) and temperature sampling ($T=1.3$, $T=1.6$). Values are reported as Prefix / $T$=1.3 / $T$=1.6. The best result in each row is shown in bold.}
\label{tab:prefix-vs-temperature}
\resizebox{\textwidth}{!}{
\begin{tabular}{ll|ccccc}
\toprule
Model & Epoch & pass@1 & pass@8 & pass@16 & pass@32 & pass@64 \\
\midrule
\multirow{4}{*}{\texttt{Qwen-2.5-0.5B}}
& 1 &
\textbf{0.36}/0.25/0.04 &
\textbf{0.73}/0.62/0.23 &
\textbf{0.81}/0.72/0.32 &
\textbf{0.88}/0.81/0.41 &
\textbf{0.92}/0.88/0.48 \\

& 2 &
\textbf{0.38}/0.31/0.02 &
\textbf{0.74}/0.68/0.12 &
\textbf{0.82}/0.78/0.18 &
\textbf{0.89}/0.85/0.25 &
\textbf{0.93}/0.91/0.29 \\

& 4 &
\textbf{0.42}/0.40/0.27 &
\textbf{0.75}/0.74/0.64 &
\textbf{0.82}/0.81/0.74 &
\textbf{0.88}/\textbf{0.88}/0.82 &
0.92/\textbf{0.93}/0.88 \\

& 8 &
\textbf{0.42}/0.41/0.35 &
\textbf{0.73}/0.72/0.71 &
\textbf{0.81}/0.80/0.80 &
\textbf{0.87}/0.86/0.86 &
\textbf{0.91}/0.90/\textbf{0.91} \\

\midrule

\multirow{4}{*}{\texttt{EvoLM-1B}}
& 1 &
\textbf{0.08}/0.05/0.02 &
\textbf{0.34}/0.26/0.13 &
\textbf{0.47}/0.39/0.20 &
\textbf{0.60}/0.53/0.27 &
\textbf{0.72}/0.66/0.33 \\

& 2 &
\textbf{0.12}/0.08/0.03 &
\textbf{0.41}/0.33/0.15 &
\textbf{0.54}/0.46/0.23 &
\textbf{0.67}/0.60/0.32 &
\textbf{0.78}/0.71/0.39 \\

& 4 &
\textbf{0.16}/0.11/0.07 &
\textbf{0.47}/0.42/0.30 &
\textbf{0.59}/0.55/0.42 &
\textbf{0.71}/0.68/0.54 &
\textbf{0.80}/0.79/0.65 \\

& 8 &
\textbf{0.17}/0.15/0.12 &
\textbf{0.48}/0.46/0.42 &
\textbf{0.59}/0.58/0.55 &
\textbf{0.70}/\textbf{0.70}/0.66 &
0.79/\textbf{0.80}/0.76 \\

\bottomrule
\end{tabular}
}
\end{table*}

\section{Relation to Prior Work on Search-Based Reasoning}
\label{app:related-sos}

Our work is closely related in spirit to recent studies arguing that exposing models to richer reasoning trajectories improves reasoning ability, particularly \textit{Stream of Search} (SoS)~\citep{gandhi2024stream} and \textit{Searchformer}~\citep{lehnert2024beyond}. These works share the intuition that training exclusively on successful reasoning traces limits the reasoning behaviors that models can acquire. However, they study different phenomena and address different research questions.

The primary focus of SoS is how a model learns to perform search \emph{within a single reasoning episode}. Working with pre-reasoning language models (i.e., before \citep{guo2025r1} and \citep{jaech2024openai}), SoS and Searchformer show that training only on successful trajectories prevents models from learning important search behaviors such as backtracking, exploring alternative branches, and recovering from mistakes. To address this limitation, they proposes training on serialized search trees that explicitly expose branching, failed attempts, and recovery steps. Searchformer similarly argues that exposing richer search trajectories improves a model's ability to perform search.

In contrast, our work investigates a phenomenon that emerges in modern \emph{post-trained reasoning models}. Rather than asking whether a model can recover from an incorrect intermediate state, we ask why repeated sampling from the same model often produces highly redundant reasoning traces despite the existence of many valid reasoning paths. We empirically observe that reasoning-oriented post-training substantially reduces reasoning-path diversity, leading to degraded $pass@k$ despite maintaining strong $pass@1$ performance. Our goal is therefore to understand why the model's sampling distribution collapses onto only a small subset of available reasoning paths.

Our proposed \emph{fork-in-the-road hypothesis} provides one possible explanation for this behavior. We argue that when training data contains \emph{indecipherable decision points}---positions where multiple valid reasoning paths or reasoning modes exist but the preceding context does not reveal why one alternative was selected---the model cannot learn a generalizable decision rule. Instead, it relies on spurious cues to resolve the ambiguity, gradually suppressing alternative reasoning trajectories during post-training. Although this hypothesis is motivated by the same high-level observation that single-trace supervision is incomplete, it addresses a different consequence of that supervision than SoS and Searchformer. Instead of explaining failures in search and recovery within a single trajectory, it explains the loss of diversity across multiple sampled trajectories.

This difference naturally leads to different evaluation protocols and interventions. Prior work evaluates a model's ability to successfully complete a search process within a single trajectory, whereas we measure the coverage of reasoning paths across repeated samples using metrics such as $pass@k$ and solution diversity. Likewise, SoS and Searchformer enrich supervision by exposing complete search trees during training, while our experiments investigate how increasing per-problem trajectory diversity during post-training and simple inference-time interventions, such as prefix manipulation, can partially recover the collapsed reasoning coverage.

\end{document}